\pdfoutput=1
\PassOptionsToPackage{sort&compress}{natbib}
\PassOptionsToPackage{hyphens}{url}
\documentclass[preprint,12pt,nopreprintline]{elsarticle}

\usepackage[T1]{fontenc}
\usepackage{graphicx}
\usepackage{booktabs}
\usepackage{amsmath,amssymb,amsfonts}
\usepackage{multirow}
\usepackage{array}
\usepackage{tabularx}
\usepackage{makecell}
\usepackage{xcolor}
\usepackage{url}
\usepackage[letterpaper,top=1in,bottom=1in,left=1in,right=1in]{geometry}
\usepackage{hyperref}
\usepackage{microtype}
\usepackage[below]{placeins}
\usepackage{adjustbox}
\usepackage{rotating}

\setkeys{Gin}{width=\linewidth,height=0.88\textheight,keepaspectratio}

\newlength{\widetabwidth}
\newlength{\widetabmax}
\newenvironment{widetab}[1][\textwidth]
  {\par\begin{adjustbox}{max width=\widetabmax,center}\begin{minipage}{#1}}
  {\end{minipage}\end{adjustbox}\par}
\AtBeginDocument{%
  \setlength{\widetabwidth}{\textwidth}%
  \setlength{\widetabmax}{\textwidth}%
}

\makeatletter
\g@addto@macro\UrlBreaks{\UrlOrds}
\makeatother

\hypersetup{
  pdftitle={Eye-Specific Supervision and Fellow-Eye Leakage Audit for Multi-Disease Binocular Fundus Recognition},
  pdfauthor={Mohammed Ekram Ullah, Abhijit Kumar Ghosh, Mohammad Zavid Parvez, Subrata Chakraborty},
  pdfkeywords={Fundus photography, ODIR-5K benchmark, Multi-label classification, Label noise, Data leakage, Probability calibration}
}

\begin{document}

\let\oldfootnoterule\footnoterule
\renewcommand{\footnoterule}{}
\begin{frontmatter}

\title{Eye-Specific Supervision and Fellow-Eye Leakage Audit for Multi-Disease Binocular Fundus Recognition}

\author[1]{Mohammed Ekram Ullah\corref{cor1}}
\ead{mohammed.ekram.ullah95@gmail.com}
\ead[url]{https://orcid.org/0009-0001-8654-1769}
\cortext[cor1]{Corresponding author}

\author[2]{Abhijit Kumar Ghosh}

\author[3]{Mohammad Zavid Parvez}

\author[4]{Subrata Chakraborty}

\address[1]{Department of Electrical and Electronic Engineering, Chittagong University of Engineering \& Technology, Pahartoli, Raozan, Chattogram 4349, Bangladesh}
\address[2]{Department of Computer Science and Engineering, BRAC University, Dhaka, Bangladesh}
\address[3]{School of Technology, Faculty of Business and Hospitality, Torrens University, Melbourne, Victoria 3000, Australia}
\address[4]{School of Science and Technology, University of New England, Armidale, NSW 2351, Australia}

\begin{abstract}
Multi-disease retinal classifiers are only as reliable as their training labels and evaluation splits, yet common practices on ODIR-5K compromise both. Although ODIR-5K pairs left and right fundus photographs with eye-specific clinician keywords, binary targets exist only at the patient level, and a popular flattened CSV drops fellow eyes while keeping just one diagnosis per eye. Most published works therefore copy patient diagnoses onto both eyes, even though nearly half of diseased patients have unilateral findings, or split the CSV randomly by image row so fellow eyes appear in both training and test sets. We audited both shortcuts and built a keyword parser that recovers 735 missing eye-level labels from the clinical text. Under a strict patient-level split across \texttt{ConvNeXt-Tiny} and \texttt{ResNet-50d}, eye-specific labels raised \texttt{ConvNeXt-Tiny} macro area under the receiver operating characteristic curve (AUC) from 0.884 to 0.911 over copied patient targets, increased hypertensive retinopathy AUC from 0.748 to 0.900, and improved zero-shot transfer to APTOS-2019. In a matched-cohort audit, random row splitting inflated macro-AUC by 1.2 percentage points from fellow-eye exposure alone. With clean eye-level supervision, parameter-free min/max pooling matched learned binocular fusion heads, and combining binary cross-entropy with Asymmetric Loss reached 0.920 macro-AUC. Post-hoc Platt scaling reduced Asymmetric Loss expected calibration error from 0.275 to 0.029. On ODIR-5K, clean eye labels, patient-level splits, and calibration matter more than complex fusion layers.
\end{abstract}


\end{frontmatter}

\clearpage
\let\footnoterule\oldfootnoterule

\section{Introduction}\label{sec:introduction}
The large and growing problem in public health is about vission loss. In year 2020, around the world 43.3 million people were blind, and 295 million had moderate or severe vision loss \cite{gbd2021trends}. Cataract, glaucoma, age-related macular degeneration (AMD), and diabetic retinopathy are among the leading causes of blindness in adults aged 50 years and older \cite{gbd2021causes}, and the number of adults with diabetic retinopathy is expected to grow from about 103 million in 2020 to 161 million by 2045 \cite{teo2021global}. The several situations and conditions can be treated if they are recognized early; often the patient observes any change in vision. This purpose fits well by the photography of color fundus. It is faster to take the image in a few seconds, non-aggressive, and cheaper than any other cost of the eye test. This color fundus image of the retina can reveal the signs of several eye diseases and also detect general health problems like hypertension. The main challenge is that taking the images is much easier than reading them, which is the bottleneck. These eye screening programs take more photos that experts can check, and also the eye specialists aren't spread widely in the world, so the areas of low-income regions face more blindness and have fewer ophthalmologists\cite{resnikoff2020estimated}.

In these circumstances, artificial intelligence (AI) and deep learning help to prevent the problem of this bottleneck. The visual sign of disease is learned by the deep network, which takes the input from the labeled images. Alongside this,the  network processes that images in a few seconds and applies the same steps for the similar type of images. Combined with telemedicine, such systems can screen patients in primary care and community settings, sort large volumes of images, and refer only the cases that need a specialist \cite{ting2019artificial}. The approach has already reached clinical use. In 2018, the first ``stand-alone'' AI diagnostic system to receive clearance from the U.S. Food and Drug Administration in any field of medicine was for diabetic retinopathy, following a prospective trial in primary care practices \cite{abramoff2018pivotal}. But therein lie two subtleties that are easy to overlook: the model must correctly label images that it learns from, and its accuracy must be evaluated on patients it has not seen before. Both of these are the focus of the current study, which is on multi-disease recognition based on a pair of fundus images.

Colour fundus photography based deep learning began with single disease detection, primarily diabetic retinopathy \cite{gulshan2016development,ting2017development}, and has progressed to multi-label models which provide multiple retinal disease labels from a single image \cite{cen2021automatic,zhou2023foundation,qiu2024development}. However, with an eye exam, it is usually not a single photo. The ophthalmologist views both the right (OD) and left (OS) images at the same time, is aware of the patient's age and sex, and compares the two optic disc images, the vessel sizes, and the macula images. To study this bilateral setting, the Ocular Disease Intelligent Recognition benchmark (ODIR-5K), released by Peking University \cite{li2021benchmark,odir2019kaggle}, was created. For each patient a pair of fundus photos is provided, and the patient is classified as one or more of the following: Normal (\texttt{N}), Diabetic Retinopathy (\texttt{D}), Glaucoma (\texttt{G}), Cataract (\texttt{C}), Age-Related Macular Degeneration (\texttt{A}), Hypertensive Retinopathy (\texttt{H}), Pathological Myopia (\texttt{M}), or Other Abnormalities (\texttt{O}).

ODIR-5K has become a standard benchmark, but there are two potential issues with the public release of ODIR-5K. The first being eye-level labels. The source spreadsheet contains free-text diagnostic keywords (one per eye), and eight binary presence/absence columns ($\texttt{N}, \texttt{D}, \texttt{G}, \texttt{C}, \texttt{A}, \texttt{H}, \texttt{M}, \texttt{O}$), which describe the \emph{patient}, not the eye. A single label file (\texttt{full\_df.csv}) is also provided, flattened. It contains $6{,}392$ eye rows from $3{,}358$ unique patients ($324$ partner eyes are missing); each eye has one class string and it is impossible to track a second disease if it occurs in the same eye. The second issue is the use of these labels and eye images in the development of a model. If the vector information for the patient is copied to both eyes ($\mathbf{y}_{\text{left}} = \mathbf{y}_{\text{right}} = \mathbf{y}_{\text{patient}}$), then a disease that is set only in one eye also gets set in the other. This situation is common in our retained cohort of $3{,}332$ patients: averaged over the seven disease categories, $43.6\%$ of positive patients have the condition in only one eye ($52.8\%$ across all positive patient-disease instances). Copying can therefore mark a fellow eye as positive for a disease recorded only in the other eye. In addition, if image rows rather than patients are split at random, one eye of a patient can be used for training while the other is used for testing.

A further question concerns the model output: one part is how to merge the two eyes into one prediction, and the other is whether its probabilities are well calibrated. Some ODIR-5K studies introduce multi-stream attention and feature-interaction modules to merge the left and right eye representations \cite{li2020dense,he2021multi,ou2022bfenet,zhao2024multilabel}; however, they rarely test how well these learned heads perform compared to simple probability pooling ($\min$ for Normal and $\max$ for disease categories) in the ODIR-5K task, given a cleanly supervised single backbone. The question of calibration arises. Asymmetric Loss (ASL) \cite{ridnik2021asymmetric} is a common choice for imbalanced multi-label data, but its focusing terms push the raw sigmoid outputs upward, so the scores have to be recalibrated before a clinical threshold is applied to them.

In this paper we evaluate the impact of three options on the outcome of ODIR-5K: (i) the construction of eye-level labels, (ii) the train--test split, and (iii) combination of the two eyes. Both classification and calibration are considered. The contributions are:
\begin{itemize}
    \item A deterministic parser covering all 94 canonical keyword phrases in ODIR-5K. It rebuilds multi-label targets for each eye and drops the pairs whose keywords describe only image artifacts.
    \item A new binocular model, \texttt{BiCA-Demo-Net}, that reads both eyes of a patient together, as an ophthalmologist does. Cross-attention links the left and right eye features, and a small demographic layer adds age and sex. The model is trained on eye-specific labels, which improve on patient-label copying on two backbones (\texttt{ConvNeXt-Tiny} and \texttt{ResNet-50d}) under a strict patient-level split.
    \item A nine-configuration audit of binocular aggregation that compares parameter-free min/max pooling, single-backbone learned fusion heads (including our bilateral cross-attention and demographic conditioning head, \texttt{BiCA-Demo-Net}), and dual-prior monocular pooling.
    \item A 3-seed matched-cohort experiment that quantifies fellow-eye split leakage by comparing row-level and patient-level splits on identical exposed and unseen test eyes.
    \item An evaluation of post-hoc Platt scaling for calibrating Asymmetric Loss probabilities, and a zero-shot test of monocular generalization on the external APTOS-2019 diabetic retinopathy cohort ($N=3{,}662$).
\end{itemize}

The paper goes as follows. Section~\ref{sec:related-work} compares with previous ODIR-5K classifiers and to label noise, data leakage, binocular fusion, and calibration studies. Section~\ref{sec:methodology} presents the label reconstruction from keywords, the retinal preprocessing, the supervision arms and fusion heads, the statistical protocol, and the metrics. In Section~\ref{sec:experiments}, the ablation, the fellow-eye leakage audit, the probability calibration, external validation with APTOS-2019, qualitative localization maps, and a comparative analysis with published ODIR-5K studies are reported. Section~\ref{sec:discussion} discusses the implications for benchmark users, as well as limitations and recommendations, and Section~\ref{sec:conclusion} provides the conclusion.

\section{Related Work}\label{sec:related-work}
We review four lines of work that shape this study: multi-disease fundus recognition on ODIR-5K, learning with noisy labels, data leakage in medical imaging, and binocular fusion with probability calibration.
\subsection{Multi-Label Fundus Recognition on ODIR-5K and Ophthalmic Foundation Models}
There are primarily two approaches to multi-disease fundus recognition since the appearance of the OIA-ODIR benchmark \cite{li2021benchmark}. In the first one, ODIR-5K is treated as a problem at the patient level, where both eyes of a patient are fed into a two-stream convolutional network or transformer network. DCNet, developed by Li et~al.\ \cite{li2020dense} and He et~al.\ \cite{he2021multi}, computes dense pixel-level and channel-level correlations between the two fundus images. The two streams are allowed to interact within BFENet \cite{ou2022bfenet}, while features are enhanced at several scales, and Zhao et~al.\ \cite{zhao2024multilabel} adopt Transformer self-attention on top of ResNet features to model interactions among disease labels. These studies split the data by patient, which is appropriate, but they supervise each monocular branch with the binary vector of the patient from the original ODIR-5K spreadsheet data. For many eyes that vector is incorrect. Averaged across all seven disease classes, $43.6\%$ of positive patients in ODIR-5K have the disease in only one eye ($52.8\%$ across all positive patient-disease instances), so the healthy fellow eye is labelled as a false positive.

The second route takes a single-eye approach by training on the flattened single-label CSV (\texttt{full\_df.csv}) that is available on Kaggle \cite{odir2019kaggle}. Islam et~al.\ \cite{islam2019source} and Wang et~al.\ \cite{wang2020multi} adapt standard CNN and EfficientNet backbones to each eye row, while Gour and Khanna \cite{gour2021multiclass} test a variety of pretrained CNN models using transfer learning. Subsets of ODIR-5K and diabetic retinopathy have been studied later, including the $K$-fold cross-validation study by Herrera-Chavez et~al.\ \cite{herrera2024multilabel} on flattened image rows, DKCNet \cite{bhati2023discriminative} (which adds squeeze-and-excitation attention to a discriminative kernel convolution block), MBSaNet \cite{wang2023combining} (which mixes convolution with self-attention), and DiaCNN \cite{shoaib2024deep} (which pairs transfer learning with custom convolutional blocks for diabetic retinopathy grading). In addition to ODIR-5K, self-supervised ophthalmic foundation models like VisionFM \cite{qiu2024development} and RETFound \cite{zhou2023foundation} have proven to be well transferable to retinal benchmarks, and large-scale clinical platforms like the 39-disease system of Cen et~al.\ \cite{cen2021automatic} have exhibited strong generalisation capabilities. When evaluating downstream benchmarks like ODIR-5K with truncated single-label rows or non-patient splits, however, reported metrics are difficult to compare between studies.

\subsection{Learning with Noisy and Imperfect Labels in Medical Imaging}
Annotation noise is a pervasive challenge in medical image analysis, arising from inter-observer variability, automated text extraction, and coarse patient-level aggregation \cite{karimi2020deep,tajbakhsh2020embracing}. The reviews by Karimi et~al.\ \cite{karimi2020deep} and Tajbakhsh et~al.\ \cite{tajbakhsh2020embracing} both note that deep networks can memorize wrong labels, and that minority classes suffer most when they do. Ju et~al.\ \cite{ju2022improving} separated two kinds of noise in medical image classification, disagreement between experts and noise from reducing a case to a single target, and handled them with dual-uncertainty estimation and a curriculum-style boosting scheme that lowers the weight of unreliable samples. Xue et~al.\ \cite{xue2022robust} filtered noisy labels with a co-training scheme in which the networks are regularized by both global and local representations. Methods of this kind are needed when the correct labels cannot be recovered. ODIR-5K is a different case. Its label noise is structured, and the correct eye-level diagnoses are still present in the clinician keyword fields (\texttt{Left-Diagnostic Keywords} and \texttt{Right-Diagnostic Keywords}). We therefore do not ask a loss function to infer which eye is diseased; we read the eye-specific multi-label vectors directly from the source clinical text.

\subsection{Data Leakage and Patient-Level Partitioning in Medical Imaging}
Splitting data below the patient level, by slice, patch, or image, is a well-known cause of optimistic results in machine-learning-based science \cite{kapoor2023leakage,roberts2021common}. Wen et~al.\ \cite{wen2020convolutional} reviewed CNN studies of Alzheimer's disease on MRI and found that more than half of the surveyed papers may have suffered from data leakage, one cause being a split made at the image level instead of the subject level. Ya{\u{g}}{\i}{\c{s}} et~al.\ \cite{yagis2021effect} measured the effect directly: when 2D slices of brain MRI were divided at random between training and test sets, test accuracy for Alzheimer's and Parkinson's disease classification was inflated by $29\%$ to $55\%$ across four datasets, because slices from one subject are nearly identical. The same effect was reported by Bussola et~al.\ \cite{bussola2021ai} in the context of digital pathology, where predictive scores were inflated up to $41\%$ when tiles of the same subject were represented in both the training and the validation sets. Roberts et~al.\ \cite{roberts2021common} included patient overlap and confounding shortcuts in a list of primary factors why early COVID-19 chest imaging models failed to stand the test in clinical practice.

This has its own variant in bilateral eye datasets: the two eyes of a patient can end up on different sides of a random divide. Even in cases of unilateral disease, the fellow eye is informative, as both eyes are photographed using the same camera and light source and share choroidal pigmentation, vessel tortuosity, and systemic influences such as age and hypertension. A number of ODIR-5K experiments \cite{islam2019source,wang2020multi,gour2021multiclass,bhati2023discriminative,herrera2024multilabel} find macro-AUC values within the range of $0.93$ to $0.97$ in random row splits of \texttt{full\_df.csv}. We measure the size of this effect with a 3-seed matched-cohort design, in which the row-split and patient-split models are scored on the same fellow-eye-exposed and unseen-patient test subsets.

\subsection{Binocular Aggregation, Class Imbalance, and Probability Calibration}
Focal Loss \cite{lin2017focal} and Asymmetric Loss (ASL) \cite{ridnik2021asymmetric} were developed to deal with imbalanced recognition. In the case of negatives, ASL employs a greater focusing exponent than in the case of positives ($\gamma_- > \gamma_+$) and changes the distribution of probabilities so that easy negatives add only a small gradient. The trade-off is probability calibration. Studies of the overall issue of calibration in computer vision \cite{guo2017calibration,nixon2019measuring} show that the scores of modern networks can vary significantly compared to observed frequencies, and a highly asymmetric objective such as ASL shifts predicted scores away from actual disease prevalence. Calibration is regained cheaply by a simple affine transformation of the logits fitted at the end of training, known as Platt scaling \cite{platt1999probabilistic}, without affecting the class ranking. On the fusion side, previously published works based on learned binocular fusion or multimodal conditioning, such as Feature-wise Linear Modulation (FiLM) \cite{perez2018film}, do not differentiate between the benefits of spatial cross-attention and the gains obtained by ensembling models trained with different losses. It is, to the best of our knowledge, the first study to bring together the four aspects analyzed here: eye-specific multi-label annotations of ODIR-5K, the decoupling of single-backbone binocular fusion and dual-prior pooling, a matched-cohort audit of fellow-eye split leakage, and post-hoc calibration alongside external monocular transfer. The topics covered above raise four questions about ODIR-5K experiments: How are the eye-level labels constructed? How are the data split? How are the two eyes combined? How well are the predicted probabilities calibrated? The next section describes how we examine these questions.

\section{Methodology}\label{sec:methodology}
\begin{figure}[!htbp]
\centering
\includegraphics[width=\linewidth]{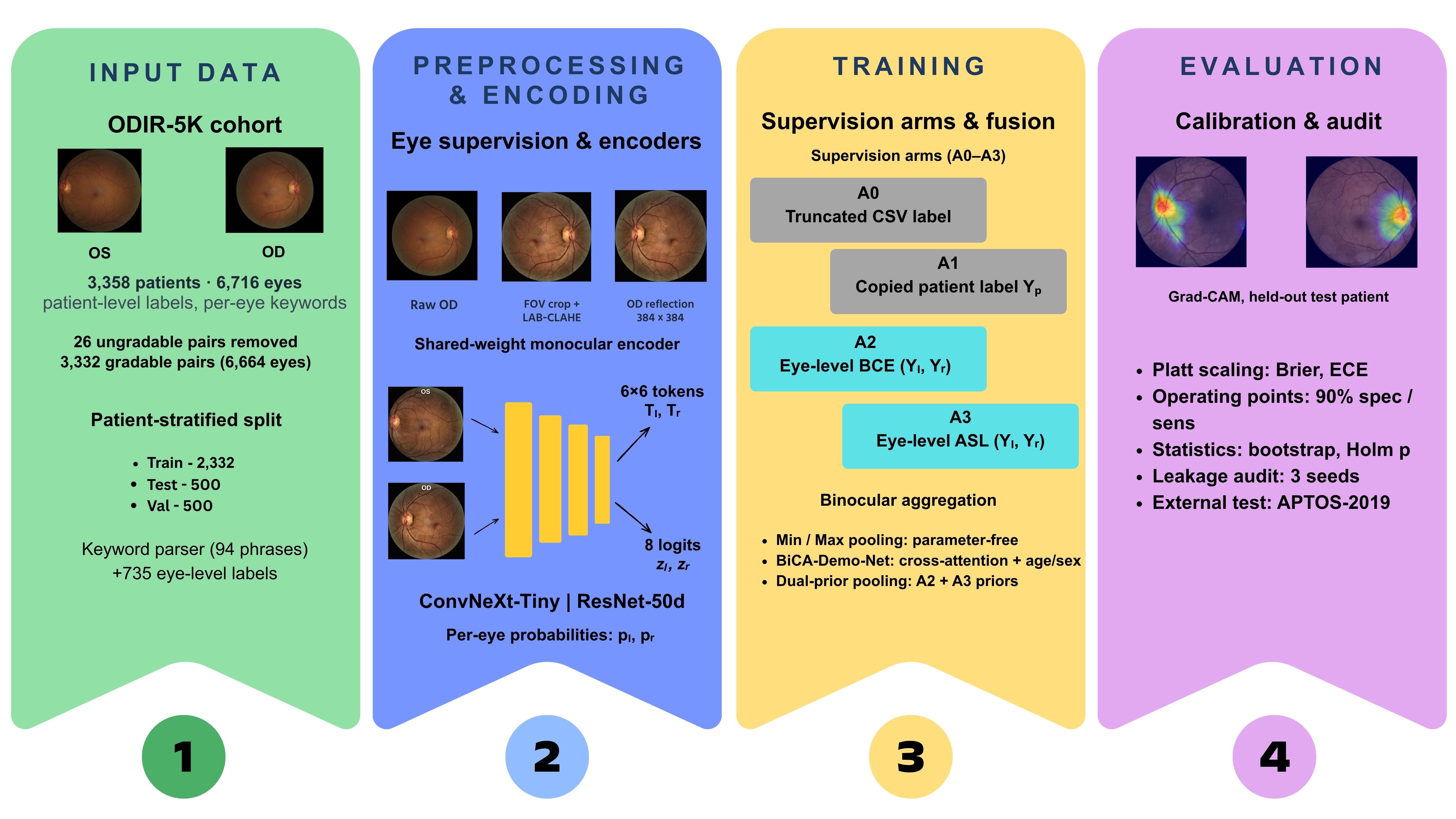}
\caption{The three-stage experimental pipeline. Stage~1: retinal preprocessing, followed by \texttt{ConvNeXt-Tiny} and \texttt{ResNet-50d} encoders that output eye-level logits and spatial feature tokens. Stage~2: training of the monocular supervision arms (\texttt{A0}--\texttt{A3}) and comparison of parameter-free logical pooling with learned binocular fusion (\texttt{BiCA-Demo-Net}) and dual-prior pooling. Stage~3: Platt scaling fitted on the validation set, evaluation at clinical operating points, paired patient-clustered bootstrap and permutation tests, and external transfer to APTOS-2019.}\label{fig:pipeline-overview}
\end{figure}

To examine these four questions, we designed the experiments shown in Figure~1. The source records are first deduplicated, and the diagnostic keywords are converted into eye-level labels. The experiments then proceed in three stages.
Figure~\ref{fig:pipeline-overview} summarizes the experimental design. Before any training, the source records are deduplicated and the 94 keyword phrases are parsed into eye-level labels; three stages follow. Stage~1 prepares the images. The left (OS) and right (OD) photographs of the patient-stratified ODIR-5K cohort ($3{,}332$ gradable patients, $6{,}664$ eyes) are normalized to a circular field of view (FOV), enhanced with contrast-limited adaptive histogram equalization (CLAHE) on the CIE-LAB luminance channel, and resized to $384\times 384$. A shared-weight monocular backbone, \texttt{ConvNeXt-Tiny} \cite{liu2022convnext} or \texttt{ResNet-50d} \cite{he2016deep,he2019bag}, then outputs eye-level logits $(z_L, z_R)$ and $6\times 6$ spatial token grids $(T_L, T_R)$; for binocular fusion, the right-eye feature map is reflected into canonical nasal--temporal orientation before the tokens are pooled. Stage~2 trains four monocular supervision arms (\texttt{A0}--\texttt{A3}) and compares nine ways of combining the two eyes, which fall into three groups: parameter-free logical pooling, learned fusion heads on a single backbone (\texttt{BiCA-Demo-Net}), and dual-prior pooling (Figure~\ref{fig:model-architecture}). In Stage~3, patient-level predictions are calibrated via validation-fitted per-class Platt scaling, evaluated at validation-locked $90\%$ specificity and sensitivity operating points with $2{,}000$-replicate patient-clustered bootstrap and Holm-corrected permutation tests, and tested zero-shot on APTOS-2019 ($N=3{,}662$) \cite{aptos2019}.

\subsection{Dataset Curation and 94-Phrase Diagnostic Taxonomy}\label{subsec:dataset-reconstruction}
To resolve the single-label truncation and missing-eye omissions of the flattened single-label CSV (\texttt{full\_df.csv}), we reconstruct the ODIR-5K dataset directly from the deduplicated patient-level records embedded in \texttt{full\_df.csv} (\texttt{groupby('ID')}, yielding $N=3{,}358$ unique patients and $6{,}716$ paired eyes consistent with OIA-ODIR metadata) \cite{li2021benchmark,odir2019kaggle}. Each patient record $i \in \{1,\dots,3358\}$ contains patient age $a_i$, biological sex $s_i \in \{0,1\}$, left and right fundus image filenames, free-text diagnostic keyword strings $(K_{i,L}, K_{i,R})$, and the 8-class patient-level ground-truth binary vector $Y_{i,P} \in \{0,1\}^8$ over $\mathcal{C} = \{\text{N, D, G, C, A, H, M, O}\}$.

Inspection of the public repository directory structure explains why $324$ eye rows were omitted from the flattened single-label CSV: the script that generated \texttt{full\_df.csv} indexed only the $6{,}392$ pre-cropped images stored inside the preprocessed image directory, skipping $324$ fellow eyes ($298$ gradable eyes and $26$ ungradable artifact eyes) stored in the raw training image directory. Most of the skipped eyes carry compound annotations in which a quality remark accompanies a clinical finding, for example ``lens dust, normal fundus'' ($n=83$), ``laser spot, moderate non proliferative retinopathy'' ($n=20$), and ``spotted membranous change'' ($n=16$). Linking both image directories resolves all $6{,}664$ gradable eyes ($6{,}366$ from the preprocessed directory and $298$ from the raw training directory: $147$ left, $151$ right) with zero missing image files.

We tokenize each eye's diagnostic keyword string by splitting on standard and full-width comma and semicolon delimiters and normalizing whitespace and case, yielding exactly $94$ canonical diagnostic phrases across the $6{,}716$-eye corpus. Rather than relying on an open-ended catch-all rule that could misclassify image-quality remarks as ocular disease, we categorize all $94$ canonical phrases into three disjoint, auditable registries (summarized in Appendix~\ref{sec:appendix}, Table~\ref{tab:appendix-taxonomy}):
\begin{enumerate}
    \item \textbf{Non-Diagnostic Artifact and Ungradable Registry (6 phrases):} ``lens dust'', ``low image quality'', ``optic disk photographically invisible'', ``anterior segment image'', ``image offset'', and ``no fundus image''. When an artifact phrase co-occurs alongside a valid clinical diagnosis (e.g., ``lens dust, normal fundus''), the artifact phrase is ignored and the clinical diagnosis is retained. When the keyword string of either eye consists \emph{exclusively} of ungradable quality remarks (the five phrases above other than ``lens dust''), the patient pair is flagged as ungradable and excluded from the study cohort ($26$ patients / $52$ eyes).
    \item \textbf{Primary Target Disease and Normal Registry (21 phrases):} These phrases map directly to Normal (\texttt{N}: ``normal fundus'') or to one of six named diseases, which we align with ICD-10 ophthalmic codes. The six are Diabetic Retinopathy (\texttt{D}, H36.0; nonproliferative and proliferative forms), Glaucoma (\texttt{G}, H40; including suspected glaucoma), Cataract (\texttt{C}, H25--H26), Age-Related Macular Degeneration (\texttt{A}, H35.3; dry, wet, and unqualified forms), Hypertensive Retinopathy (\texttt{H}, H35.0), and Pathological Myopia (\texttt{M}, H44.2; including myopic maculopathy and myopia retinopathy).
    \item \textbf{Verified ``Other Abnormality'' Registry (67 phrases):} The remaining $67$ canonical ophthalmic pathology phrases (such as branch and central retinal vein occlusion, epiretinal membrane, macular hole, drusen, chorioretinal atrophy, retinitis pigmentosa, optic disc edema, and vitreous degeneration) are explicitly listed and mapped to class \texttt{O}. Any remaining unmapped phrase set is explicitly verified to be empty ($0$ unmapped phrases) across all $6{,}716$ eyes.
\end{enumerate}

For every patient $i$, let $Y_{i,L}, Y_{i,R} \in \{0,1\}^8$ denote the resulting eye-specific multi-label vectors. By the clinical definition of the ODIR-5K patient-level ground truth, the reconstructed patient vector $\hat{Y}_{i,P}$ is given by the boolean aggregation:
\begin{equation}\label{eq:logical-or}
\begin{aligned}
\hat{Y}_{i,P,c} &= \max\!\left(Y_{i,L,c},\, Y_{i,R,c}\right) \quad \forall c \in \{\text{D, G, C, A, H, M, O}\}, \\
\hat{Y}_{i,P,N} &= \mathbb{I}\!\left(\sum_{c \neq N} \hat{Y}_{i,P,c} = 0 \;\wedge\; \max\!\left(Y_{i,L,N},\, Y_{i,R,N}\right) = 1\right).
\end{aligned}
\end{equation}
Evaluating Eq.~\eqref{eq:logical-or} across all $3{,}358$ source patients yields $3{,}358/3{,}358$ ($100.00\%$) exact vector equality ($\hat{Y}_{i,P} \equiv Y_{i,P}$). This $100.00\%$ agreement confirms deterministic \emph{internal logical consistency} between our eye-level keyword parser and the source patient-level annotations, rather than independent clinical re-grading of the raw fundus photographs. After removing the $26$ ungradable patient pairs, the retained study cohort comprises $N=3{,}332$ paired binocular patients ($6{,}664$ eyes), partitioned via rarity-prioritized patient-level stratification into $2{,}332$ training patients ($4{,}664$ eyes, $70\%$), $500$ validation patients ($1{,}000$ eyes, $15\%$), and $500$ locked test patients ($1{,}000$ eyes, $15\%$) with zero patient overlap across splits (Table~\ref{tab:dataset-flow}).

\begin{table}[!htbp]
\centering
\caption{Cohort construction and quality control, from the public ODIR-5K repository to the final patient-independent partitions (first 16 characters of the SHA-256 manifest hash: \texttt{2307897db1f1212e}).}\label{tab:dataset-flow}
\scriptsize
\setlength{\tabcolsep}{5pt}
\renewcommand{\arraystretch}{1.15}
\begin{widetab}[\textwidth]
\begin{tabularx}{\linewidth}{@{}lrrX@{}}
\toprule
\textbf{Cohort Curation Stage} & \textbf{Patients} & \textbf{Eyes} & \textbf{Criteria \& Provenance Notes} \\
\midrule
1. Source Repository (Deduplicated) & $3{,}358$ & $6{,}716$ & $6{,}392$ flattened CSV rows + $324$ unindexed fellow eyes ($100.0\%$ internal consistency with source patient vectors) \\
2. Artifact-Only Exclusions & $-26$ & $-52$ & Ungradable quality remarks only (``low image quality'', ``optic disk photographically invisible'', ``anterior segment image'', ``no fundus image'') \\
3. Missing Image Exclusions & $0$ & $0$ & All image files found across the preprocessed ($6{,}366$ eyes) and raw training ($298$ eyes) directories \\
\midrule
\textbf{4. Retained Clean Gradable Cohort} & $\mathbf{3{,}332}$ & $\mathbf{6{,}664}$ & \textbf{Every patient has both eyes, each with a verified eye-specific multi-label target} \\
\hspace{0.8em}-- Training Partition ($70\%$) & $2{,}332$ & $4{,}664$ & No patient shared with validation or test; train age $57.98 \pm 11.64$~yr \\
\hspace{0.8em}-- Validation Partition ($15\%$) & $500$ & $1{,}000$ & Early stopping, threshold selection, and Platt scaling fits \\
\hspace{0.8em}-- Locked Test Partition ($15\%$) & $500$ & $1{,}000$ & Scored once for each model configuration \\
\bottomrule
\end{tabularx}
\end{widetab}
\end{table}

\subsection{Retinal Preprocessing and Canonical Nasal--Temporal Alignment}\label{subsec:preprocessing}
\begin{figure}[!htbp]
\centering
\includegraphics[width=\linewidth]{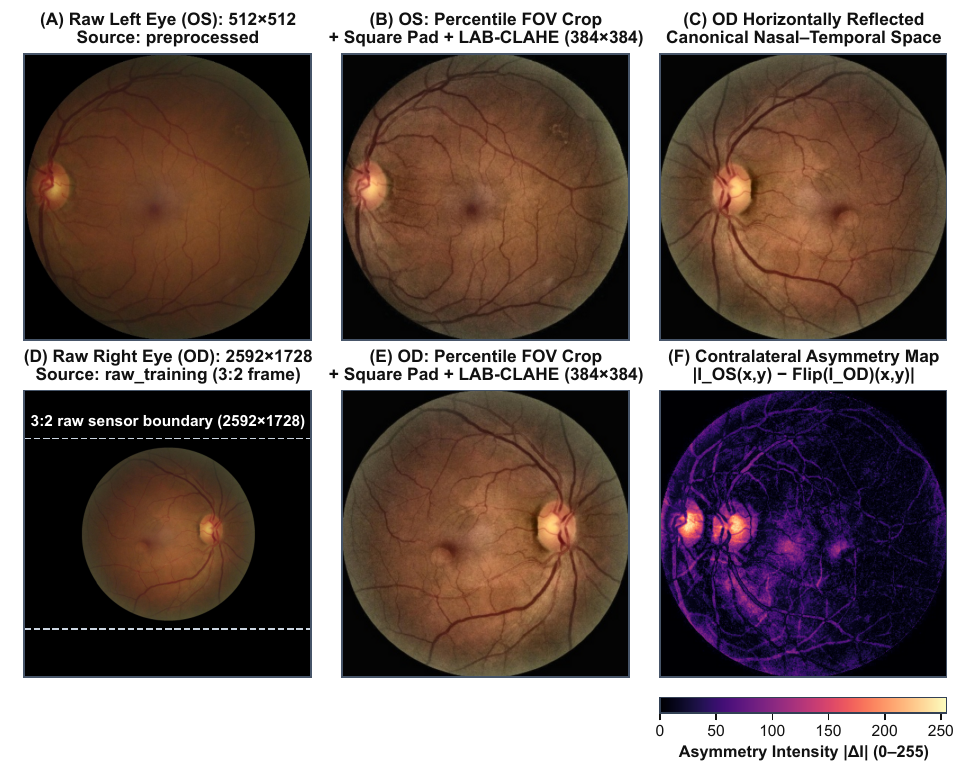}
\caption{Unified circular FOV normalization and canonical nasal--temporal alignment displayed in uniform $1{:}1$ square frames (Patient ID=25, Age=63; internal provenance cache key \texttt{otsu\_\allowbreak fov\_\allowbreak sqpad\_\allowbreak lab\_\allowbreak clahe\_\allowbreak 384\_\allowbreak v4.1}). (A, D) Raw left (OS, $512\times 512$) and right (OD, $2592\times 1728$ centered within a $1{:}1$ frame showing its native $3{:}2$ sensor boundary) fundus photographs from the two repository directories. (B, E) Standardized $384\times 384$ images after percentile-floored intensity FOV cropping ($\max(12, P_5(I_{\text{gray}}))$), aspect-ratio-preserving square padding, and CIE-LAB luminance CLAHE. (C) Canonical horizontal reflection aligning right-eye nasal--temporal coordinates with OS. (F) Pixel-wise contralateral asymmetry map $|I_{\text{OS}}(x,y) - \text{Flip}(I_{\text{OD}})(x,y)|$.}\label{fig:preprocessing}
\end{figure}

The public ODIR-5K files are not uniform: $6{,}366$ eyes are already cropped to the circular fundus, while $298$ are raw rectangular camera frames with wide black borders. We therefore pass every image, from ODIR-5K and from the external APTOS-2019 set alike, through the same preprocessing steps (Figure~\ref{fig:preprocessing}):
\begin{enumerate}
    \item \textbf{Intensity-Thresholded Circular FOV Cropping and Square Padding:} For an RGB input $I$ with grayscale intensity $I_{\text{gray}}$, the foreground mask is $M(y,x) = \mathbb{I}(I_{\text{gray}}(y,x) > \max(12, P_5(I_{\text{gray}})))$. The image is cropped to the bounding box $[y_0:y_1, x_0:x_1]$ of this mask, which is the circular retinal FOV. To prevent aspect-ratio distortion of the optic disc cup-to-disc ratio, unequal height and width $(H, W)$ are symmetrically zero-padded to $S = \max(H,W)$ before area-interpolated resizing to $384\times 384$ pixels.
    \item \textbf{Luminance-Channel CLAHE Enhancement:} Illumination varies between cameras, and small microvascular lesions (microaneurysms, cotton-wool spots, arteriolar narrowing) are easily lost in dark regions. We map the resized image from RGB to CIE-LAB, equalize only the luminance channel $L^*$ with CLAHE (clip limit $2.0$, $8\times 8$ tile grid), and map the result back to RGB.
    \item \textbf{Canonical Nasal--Temporal Reflection and Synchronized Augmentation:} The left (OS) and right (OD) eyes are mirror images of each other along the horizontal axis, so the optic disc, which sits on the nasal side, appears at opposite horizontal positions in a left-eye and a right-eye photograph. Accordingly, monocular encoders (\texttt{A0}--\texttt{A3}) process the $384\times 384$ unreflected OS and OD frames with random geometric and photometric augmentations (horizontal and vertical flips, $\pm 15^\circ$ affine rotations, $\pm 4\%$ translation, $0.95\text{--}1.05\times$ scaling, and color jitter) synchronized across both eyes of a patient via a shared random seed. For downstream binocular feature comparison (arms \texttt{A4}--\texttt{A6}), the right-eye spatial feature map is horizontally reflected into canonical nasal--temporal space prior to $6\times 6$ adaptive spatial pooling (Figure~\ref{fig:preprocessing}C, F).
\end{enumerate}

\subsection{Supervision Arms and Multi-Label Loss Priors}\label{subsec:losses}
Label quality and loss function are varied separately across four monocular supervision arms (\texttt{A0}--\texttt{A3}). Each arm trains a shared-weight feature extractor $f_\theta$: \texttt{ConvNeXt-Tiny} \cite{liu2022convnext} at $384\times 384$ (feature dimension $D=768$), with \texttt{ResNet-50d} \cite{he2016deep,he2019bag} ($D=2048$) as the replication backbone. The arms are:
\begin{itemize}
    \item \textbf{\texttt{A0} (Flattened Single-Label CSV Baseline, BCE):} Trained on eye-level targets where multi-disease eyes in the training set are truncated to a single label using the priority order $[\text{D, G, C, A, M, H, O, N}]$ so that all $4{,}664$ paired training eyes (including the $215$ training eyes omitted from \texttt{full\_df.csv}) receive a single label on the exact same cohort as \texttt{A1}--\texttt{A6}. Across the $4{,}449$ training eyes present in the flattened single-label CSV, this priority rule matches the raw single-label column in $97.68\%$ ($4{,}346/4{,}449$) of eyes; for Class \texttt{H}, the raw CSV retains $87$ positive training eyes whereas priority truncation retains $75$ ($12$ fewer out of $136$ true Class-\texttt{H} training eyes, $135$ of which are present in \texttt{full\_df.csv}), a minor difference noted in Section~\ref{subsec:limitations}. The loss is unweighted binary cross-entropy (\texttt{BCE}).
    \item \textbf{\texttt{A1} (Copied Patient-to-Eye Baseline, BCE):} Trained by assigning the patient-level target vector $Y_{i,P}$ to both eyes ($\mathcal{L} = \frac{1}{2}[\ell_{\text{BCE}}(z_{i,L}, Y_{i,P}) + \ell_{\text{BCE}}(z_{i,R}, Y_{i,P})]$) under unweighted \texttt{BCE}, replicating standard patient-label copying.
    \item \textbf{\texttt{A2} (Eye-Specific Supervision, BCE):} Trained on our keyword-reconstructed eye-specific multi-label targets $(Y_{i,L}, Y_{i,R})$ using unweighted Binary Cross-Entropy (\texttt{BCE}, Eq.~\eqref{eq:clean-eye-loss}):
    \begin{equation}\label{eq:clean-eye-loss}
    \mathcal{L}_{\text{eye}}(\theta) = \frac{1}{2}\Big[\ell_{\text{BCE}}\!\left(z_{i,L},\, Y_{i,L}\right) + \ell_{\text{BCE}}\!\left(z_{i,R},\, Y_{i,R}\right)\Big],
    \end{equation}
    isolating the pure benefit of eye-specific multi-label supervision relative to \texttt{A0} and \texttt{A1} without class reweighting.
    \item \textbf{\texttt{A3} (Eye-Specific Supervision, ASL):} Trained on $(Y_{i,L}, Y_{i,R})$ using Asymmetric Loss (\texttt{ASL}, Eq.~\eqref{eq:asl}) \cite{ridnik2021asymmetric} with positive focusing parameter $\gamma_+ = 0$, negative focusing parameter $\gamma_- = 4$, and probability margin $m = 0.05$:
    \begin{equation}\label{eq:asl}
    \begin{aligned}
    \ell_{\text{ASL}}(z, y) &= -\frac{1}{|\mathcal{C}|}\sum_{c=1}^{|\mathcal{C}|} \Big[\, y_c (1 - p_c)^{\gamma_+} \log(p_c) \\
    &\qquad\qquad + (1 - y_c)\big(p_{m,c}\big)^{\gamma_-}\log(1 - p_{m,c}) \,\Big], \\
    \text{where}\quad p_c &= \sigma(z_c), \qquad p_{m,c} = \max(p_c - m,\, 0).
    \end{aligned}
    \end{equation}
\end{itemize}
For all monocular arms (\texttt{A0}--\texttt{A3}), eye-level probabilities $p_{i,L} = \sigma(z_{i,L})$ and $p_{i,R} = \sigma(z_{i,R})$ are combined into patient-level predictions $p_{i,\text{pool}} \in [0,1]^8$ via parameter-free min/max probability pooling mirroring Eq.~\eqref{eq:logical-or}:
\begin{equation}\label{eq:prob-pool}
\begin{aligned}
p_{i,\text{pool},N} &= \min\!\left(p_{i,L,N},\, p_{i,R,N}\right), \\
p_{i,\text{pool},c} &= \max\!\left(p_{i,L,c},\, p_{i,R,c}\right) \quad \forall c \in \{\text{D, G, C, A, H, M, O}\}.
\end{aligned}
\end{equation}

\subsection{Binocular Aggregation and \texttt{BiCA-Demo-Net} Architecture}\label{subsec:bica}
\begin{figure}[!htbp]
\centering
\includegraphics[width=\linewidth,height=0.72\textheight,keepaspectratio]{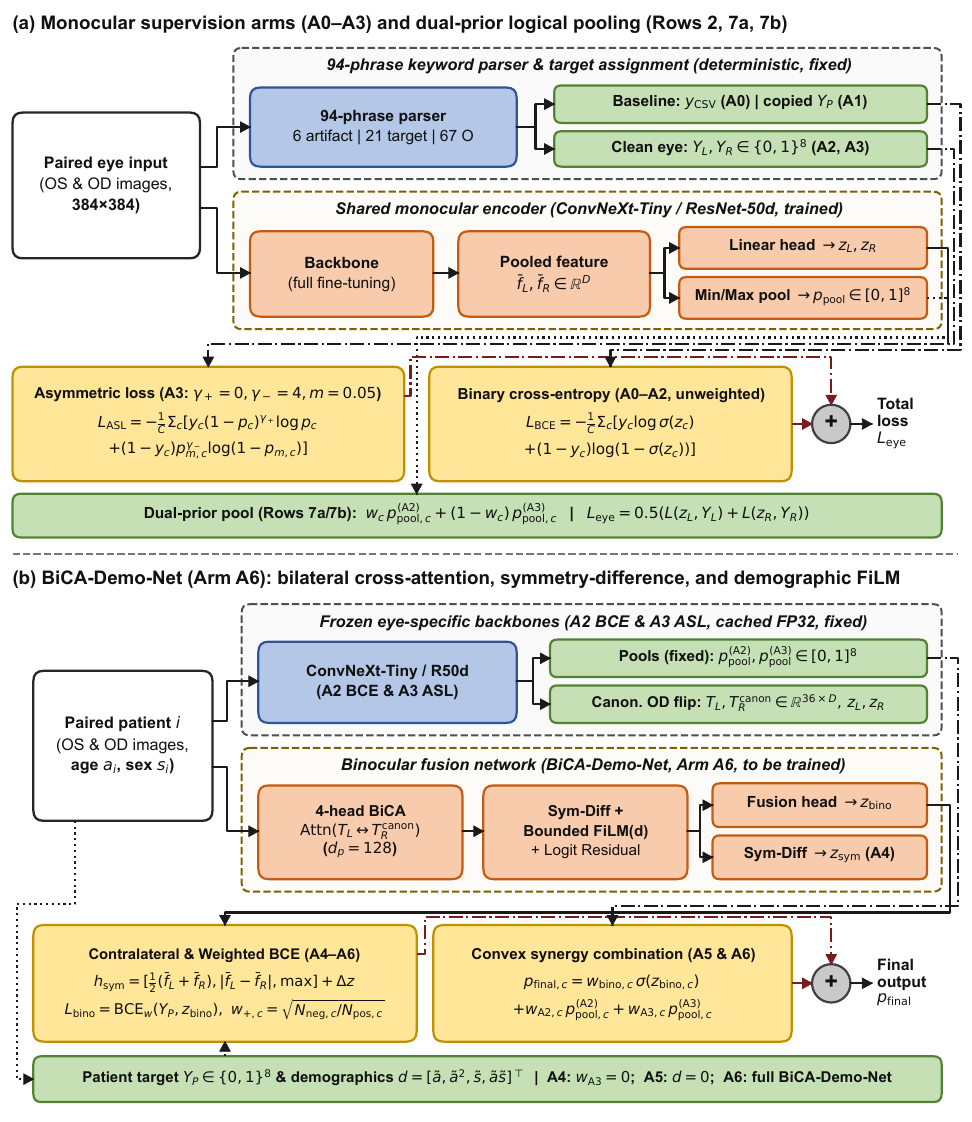}
\caption{Detailed architecture of the monocular supervision arms and binocular aggregation modules. (a)~Monocular supervision arms (\texttt{A0}--\texttt{A3}) comparing truncated CSV (\texttt{A0}) and copied patient targets (\texttt{A1}) against keyword-reconstructed eye-specific multi-label supervision under unweighted BCE (\texttt{A2}) and Asymmetric Loss (\texttt{A3}, $\gamma_+=0, \gamma_-=4, m=0.05$), alongside parameter-free min/max and dual-prior logical pooling (Rows~2, 7a, 7b). (b)~Proposed \texttt{BiCA-Demo-Net} (arm \texttt{A6}) combining 4-head bilateral cross-attention over $6\times 6$ canonical nasal--temporal spatial tokens, contralateral symmetry-difference features (\texttt{A4}), bounded demographic Feature-wise Linear Modulation (FiLM), eye-logit interaction and demographic residuals, weighted patient-level BCE training, and per-class weighted synergy combination with cached \texttt{A2} and \texttt{A3} monocular pools.}\label{fig:model-architecture}
\end{figure}

To test whether learned bilateral feature fusion improves upon parameter-free min/max pooling (Eq.~\eqref{eq:prob-pool}), we horizontally reflect the right-eye spatial feature map $F_R$ into canonical nasal--temporal space ($\text{Flip}_W(F_R)$), extract adaptively pooled $6\times 6$ feature token grids $T_L, T_R \in \mathbb{R}^{36 \times D}$ and global mean-pooled eye embeddings $\bar{f}_L, \bar{f}_R \in \mathbb{R}^D$, and cache the corresponding eye-level logits $(z_L, z_R)$ from the frozen \texttt{A2} backbone (Figure~\ref{fig:model-architecture}b). In our bilateral cross-attention and demographic fusion head (\texttt{BiCA-Demo-Net}, arm \texttt{A6}), left and right spatial tokens are projected to $d_p = 128$ dimensions and processed via 4-head bidirectional cross-attention with residual token pooling. The resulting cross-attended summary is concatenated with global symmetry-difference features computed over the pooled embeddings ($[0.5(\bar{f}_L+\bar{f}_R),\, |\bar{f}_L-\bar{f}_R|,\, \max(\bar{f}_L,\bar{f}_R)]$, evaluated standalone as arm \texttt{A4}) to form a fused binocular representation (evaluated without demographics as arm \texttt{A5}). In \texttt{BiCA-Demo-Net} (\texttt{A6}), a 4-dimensional demographic vector $d = [\tilde{a},\, \tilde{a}^2,\, \tilde{s},\, \tilde{a}\tilde{s}]^\top$ constructed from standardized age $\tilde{a} = (a - \mu_a)/\sigma_a$ and centered sex $\tilde{s} = s - 0.5$ modulates the fused representation via a bounded Feature-wise Linear Modulation (FiLM) layer \cite{perez2018film} ($h_{\text{fuse}} \leftarrow h_{\text{fuse}} \odot (1 + 0.3\tanh(\gamma(d))) + 0.3\beta(d)$). The final fusion head combines this modulated representation with a residual eye-logit interaction MLP ($32 \to 64 \to 8$, scaled by $0.35$) acting on $[\min(z_L,z_R),\, \max(z_L,z_R),\, 0.5(z_L+z_R),\, |z_L-z_R|]$ and an additive demographic prior ($4 \to 16 \to 8$, scaled by $0.25$) to yield the binocular fusion logits $z_{\text{bino}} \in \mathbb{R}^8$.

Unlike the monocular arms (\texttt{A0}--\texttt{A3}), which use unweighted losses, all second-stage token-cached binocular fusion heads (\texttt{A4}--\texttt{A6}, Rows~3--6 of Table~\ref{tab:fusion-comparison}, and \texttt{R50d\_A6}) are trained against patient-level targets $Y_{i,P}$ using damped inverse-prevalence weighted Binary Cross-Entropy:
\begin{equation}\label{eq:fusion-weighted-bce}
\begin{aligned}
\mathcal{L}_{\text{bino}} &= \frac{1}{|\mathcal{C}|}\sum_{c=1}^{|\mathcal{C}|} \Big[\, -w_{+,c}\, Y_{i,P,c}\log\sigma(z_{\text{bino},i,c}) \\
&\qquad\qquad - (1 - Y_{i,P,c})\log\big(1 - \sigma(z_{\text{bino},i,c})\big) \,\Big], \\
w_{+,c} &= \sqrt{\frac{N_{\text{train}} - N_{\text{pos},c}}{\max(N_{\text{pos},c}, 1)}},
\end{aligned}
\end{equation}
where $N_{\text{train}} = 2{,}332$ and $N_{\text{pos},c}$ denote the total patient count and patient-level positive count for class $c$ in the training split, dampening extreme inverse-frequency weights on rare classes. Validation checkpoint selection monitors the blended validation Macro-AUC of $0.35\,\sigma(z_{\text{bino}}) + 0.65\,p_{\text{pool}}^{(\text{A2})}$.

Because unweighted \texttt{BCE} (\texttt{A2}) and \texttt{ASL} (\texttt{A3}) exhibit complementary class-wise strengths, the patient-level probability for class $c$ in arms \texttt{A5} and \texttt{A6} (\texttt{BiCA-Demo-Net}) is formed by a per-class weighted combination (Eq.~\eqref{eq:synergy}) whose weights are chosen by grid search to maximize the per-class validation ROC-AUC ($w_{\text{bino},c} \in \{0.00, 0.05, \dots, 0.70\}$, $w_{\text{A3},c} \in \{0.00, 0.05, \dots, 0.50\}$, $w_{\text{A2},c} = \max(0, 1 - w_{\text{bino},c} - w_{\text{A3},c})$) on the validation set:
\begin{equation}\label{eq:synergy}
\begin{aligned}
p_{\text{final}, c} &= w_{\text{bino}, c}\,\sigma(z_{\text{bino}, c}) + w_{\text{A2}, c}\,p_{\text{pool}, c}^{(\text{A2})} + w_{\text{A3}, c}\,p_{\text{pool}, c}^{(\text{A3})}, \\
\text{with}\quad w_{\text{A2},c} &= \max\!\big(0,\; 1 - w_{\text{bino},c} - w_{\text{A3},c}\big), \qquad w_{\cdot,c} \ge 0.
\end{aligned}
\end{equation}
The three weights sum to one whenever $w_{\text{bino},c} + w_{\text{A3},c} \le 1$, which holds for every class in the selected solution, so the final combination is convex. In the 7-arm progression (Table~\ref{tab:master-ablation}), \texttt{A4} blends the symmetry-difference head with \texttt{A2} only ($w_{\text{A3},c}=0$), whereas \texttt{A5} and \texttt{A6} (\texttt{BiCA-Demo-Net}) allow $w_{\text{A3},c} \ge 0$. To separate the effect of the neural fusion head ($w_{\text{bino},c} > 0$) from the dual-prior ensemble of \texttt{A2} and \texttt{A3}, we perform a 9-row binocular aggregation ablation (Table~\ref{tab:fusion-comparison}, Section~\ref{subsec:fusion-results}) that evaluates six single-backbone aggregation methods on \texttt{A2} alone ($w_{\text{A3},c} = 0$: mean pooling, min/max logical pooling, feature concatenation MLP, symmetry-difference MLP, cross-attention without demographics, and \texttt{BiCA-Demo-Net}) against fixed $50/50$ dual-prior pooling ($0.5\,p_{\text{pool}}^{(\text{A2})} + 0.5\,p_{\text{pool}}^{(\text{A3})}$, Row 7a), validation-weighted dual-prior pooling ($w_c\,p_{\text{pool},c}^{(\text{A2})} + (1-w_c)\,p_{\text{pool},c}^{(\text{A3})}$, Row 7b), and the full dual-prior \texttt{BiCA-Demo-Net} system (Row 8).

\subsection{Matched-Cohort Fellow-Eye Leakage Audit, Platt Scaling, and Statistical Protocol}\label{subsec:stats-protocol}
\paragraph{Multi-Seed Matched-Cohort Leakage Protocol.}
In a random 70/15/15 eye-row split of $6{,}664$ eyes, the fellow eye of a test eye falls in the training set with a probability of about $0.70$. Our matched-cohort design fixes this exposure at $362$ of the $500$ test patients ($72.4\%$). Comparing a row-split model on its random test set against a patient-split model on a different test set confounds fellow-eye exposure with test-cohort composition. To eliminate this confound, for each random seed $s \in \{42, 43, 44\}$, we partition the $500$ locked test patients via stratified sampling into $362$ \emph{Fellow-Eye-Exposed} patients ($72.4\%$) and $138$ \emph{Unseen-Patient} patients ($27.6\%$, $276$ eyes). For the Row-Split Model (the same \texttt{ConvNeXt-Tiny} backbone with a standard linear classification head, trained on single-eye batches with unweighted \texttt{BCE} as in \texttt{A2}), the $362$ left eyes of the exposed test patients are swapped into the training set (replacing $362$ stratified left eyes from the training partition, keeping total training size fixed at $4{,}664$ eyes), while their $362$ right eyes are held out for testing alongside the $276$ eyes ($138$ patients) with neither eye in training. Both the Row-Split Model and the frozen Patient-Split Model (\texttt{A2}) are evaluated on the \emph{exact same} $362$ exposed test eyes ($\Delta_{\text{Exposed}} = \text{AUC}_{\text{Row}} - \text{AUC}_{\text{Pat}}$) and the \emph{exact same} $276$ unseen-patient test eyes ($\Delta_{\text{Unseen}} = \text{AUC}_{\text{Row}} - \text{AUC}_{\text{Pat}}$). We also report the within-seed \emph{Difference-in-Differences} ($\text{DiD} = \Delta_{\text{Exposed}} - \Delta_{\text{Unseen}}$), which cancels out any baseline training-schedule difference between the two models and isolates the fellow-eye exposure effect.

\paragraph{Validation-Fitted Platt Scaling and Clinical Operating Points.}
To correct post-hoc probability distortion without using test labels, we convert clamped probabilities ($\text{clip}(\hat{p}, 10^{-6}, 1-10^{-6})$) to logits $z_{i,c}$ and apply per-class affine-logit \textbf{Platt scaling} (Eq.~\eqref{eq:platt}) \cite{platt1999probabilistic,guo2017calibration}:
\begin{equation}\label{eq:platt}
z'_{i,c} = \frac{z_{i,c}}{T_c} + b_c, \qquad \hat{p}^{\text{cal}}_{i,c} = \sigma(z'_{i,c}),
\end{equation}
fitting the scale $T_c \in [0.3, 3.5]$ ($33$ uniform grid points) and bias $b_c \in [-2.5, 1.0]$ ($36$ uniform grid points) per class by minimizing the Brier score exclusively on the validation set. Similarly, for clinical screening operating points, we determine on the validation set the per-class thresholds $\tau_{90\text{Spec}, c}$ and $\tau_{90\text{Sens}, c}$ that maximize validation sensitivity subject to $\text{Spec}_{\text{val},c} \ge 0.90$ and maximize validation specificity subject to $\text{Sens}_{\text{val},c} \ge 0.90$, lock those thresholds, and evaluate the achieved test sensitivity and specificity on the held-out test set.

\paragraph{Patient-Clustered Bootstrap and Permutation Testing.}
All uncertainty intervals on ODIR-5K are estimated via $B=2{,}000$ patient-clustered bootstrap resamples, sampling unique patient IDs with replacement and retaining both correlated eyes within each sampled patient cluster. For paired model comparisons, we report $95\%$ patient-clustered bootstrap confidence intervals alongside two-sided $p$-values from a Monte Carlo paired patient-clustered sign-flip permutation test ($2{,}000$ random within-patient model-prediction swaps with the Phipson--Smyth $(k+1)/(B+1)$ correction \cite{phipson2010permutation} and Holm--Bonferroni step-down adjustment \cite{holm1979simple}).

\subsection{Implementation Details and Training Protocol}\label{subsec:implementation-details}
We implemented all models in PyTorch with the \texttt{timm} library and trained them with FP16 mixed precision on one NVIDIA Tesla T4 GPU ($16$~GB VRAM). Both backbones were initialized from ImageNet-pretrained \texttt{timm} weights (\texttt{convnext\_tiny.\allowbreak fb\_in22k\_\allowbreak ft\_in1k\_384} and \texttt{resnet50d}) and fine-tuned end to end. For the monocular arms (\texttt{A0}--\texttt{A3}), \texttt{ConvNeXt-Tiny} ($D=768$) was trained for $15$ epochs and \texttt{ResNet-50d} ($D=2048$) was trained for $12$ epochs at $384\times 384$ resolution using \texttt{AdamW} (initial learning rate $3\times 10^{-4}$, weight decay $10^{-2}$), cosine annealing ($\eta_{\min} = 10^{-6}$), a physical batch size of $16$ patient pairs ($32$ eyes) with $2$-step gradient accumulation (effective batch size $32$ patient pairs), dropout $p=0.30$, and $L_2$ gradient clipping at $1.0$. The checkpoint achieving the highest validation 8-class Macro-AUC ($N=500$; training age $\mu_a = 57.98$, $\sigma_a = 11.64$~years) was selected for testing. For the second-stage fusion heads (\texttt{A4}--\texttt{A6} and single-backbone variants in Table~\ref{tab:fusion-comparison}), spatially pooled $6\times 6$ feature tokens and eye logits were cached in FP32 from the frozen best checkpoints, and the fusion head ($d_p=128$, 4 attention heads, dropout $0.35$) was trained for $45$ epochs using \texttt{AdamW} (learning rate $10^{-3}$, weight decay $2\times 10^{-2}$, cosine annealing $\eta_{\min} = 10^{-5}$, batch size $64$, with $50\%$ random left--right token swapping and Gaussian feature noise $\sigma_{\text{tok}}=0.05, \sigma_{\text{logit}}=0.15$) under the patient-level weighted BCE loss in Eq.~\eqref{eq:fusion-weighted-bce}. The main ablation arms (\texttt{A0}--\texttt{A6}) were trained under a fixed global seed ($\text{seed} = 42$), while the matched-cohort leakage audit was repeated across three seeds ($\text{seeds} \in \{42, 43, 44\}$, $12$ epochs per seed, batch size $32$ eyes). Exact software identifiers are listed in Appendix~\ref{sec:appendix}.

\subsection{Evaluation Metrics}\label{subsec:metrics}
To evaluate multi-disease ranking, thresholded classification, and probability calibration across all $N_{\text{test}}=500$ held-out test patients and $C=8$ diagnostic classes, we report the following metrics:
\begin{itemize}
    \item \textbf{8-Class Macro-AUC and Minority-AUC:} $\text{Macro-AUC}$ averages the area under the ROC curve over the eight classes ($\mathcal{C} = \{\text{N, D, G, C,}\allowbreak\ \text{A, H, M, O}\}$) with equal weight. $\text{Minority-AUC}$ does the same over the three rarest named diseases, Age-Related Macular Degeneration (\texttt{A}, $4.2\%$ eye prevalence), Hypertensive Retinopathy (\texttt{H}, $2.9\%$), and Pathological Myopia (\texttt{M}, $3.9\%$), so that performance on low-prevalence conditions is visible separately.
    \item \textbf{Macro Mean Average Precision (mAP):} Per-class Average Precision, i.e., the area under the precision--recall curve, averaged over the eight classes with equal weight.
    \item \textbf{Micro F1, Cohen's Kappa ($\kappa$), and Official ODIR Score:} Following the OIA-ODIR benchmark protocol \cite{li2021benchmark}, predicted patient probabilities are binarized using per-class validation F1-optimal thresholds $\tau_{\text{F1},c} \in [0.10, 0.90]$ ($33$ grid points, step $0.025$) selected on the validation set, and flattened across all $N_{\text{test}} \times C$ binary decisions to compute micro-averaged $\text{F1}$ and Cohen's $\kappa$. The composite official competition metric (Eq.~\eqref{eq:odir-score}) is the unweighted mean of $\kappa$, $\text{F1}$, and $\text{Macro-AUC}$:
    \begin{equation}\label{eq:odir-score}
    \text{ODIR Score} = \frac{\kappa + \text{F1} + \text{Macro-AUC}}{3}.
    \end{equation}
    \item \textbf{Brier Score and Expected Calibration Error (ECE):} Multi-label probability accuracy is quantified by the macro-averaged Brier score and the 10-bin uniform Expected Calibration Error \cite{guo2017calibration,nixon2019measuring}:
    \begin{gather*}
    \text{Brier} = \frac{1}{C N_{\text{test}}}\sum_{c=1}^{C}\sum_{i=1}^{N_{\text{test}}}(\hat{p}_{i,c} - Y_{i,P,c})^2, \\
    \text{ECE} = \frac{1}{C}\sum_{c=1}^{C}\sum_{b=1}^{10}\frac{|B_{c,b}|}{N_{\text{test}}}\big|\text{acc}(B_{c,b}) - \text{conf}(B_{c,b})\big|.
    \end{gather*}
\end{itemize}

\section{Experiments and Results}\label{sec:experiments}

Results on ODIR-5K and on the external APTOS-2019 set are organized in seven parts: (i) label recovery and contralateral asymmetry across the dataset (Section~\ref{subsec:results-dataset}), (ii) eye-specific supervision and cross-backbone ablation on \texttt{ConvNeXt-Tiny} and \texttt{ResNet-50d} (Section~\ref{subsec:results-supervision}), (iii) single-backbone binocular fusion versus dual-prior pooling in the 9-row aggregation audit (Section~\ref{subsec:fusion-results}), (iv) the 3-seed matched-cohort fellow-eye split leakage audit (Section~\ref{subsec:leakage-results}), (v) Platt scaling and validation-locked clinical operating points (Section~\ref{subsec:calibration-results}), (vi) zero-shot monocular transfer to APTOS-2019 and bilateral Grad-CAM localization (Section~\ref{subsec:aptos-results}), and (vii) comparative analysis against published ODIR-5K studies (Section~\ref{subsec:comparative-results}).

\subsection{Dataset Label Recovery and Contralateral Asymmetry}\label{subsec:results-dataset}
\begin{figure}[!htbp]
\centering
\includegraphics[width=\linewidth]{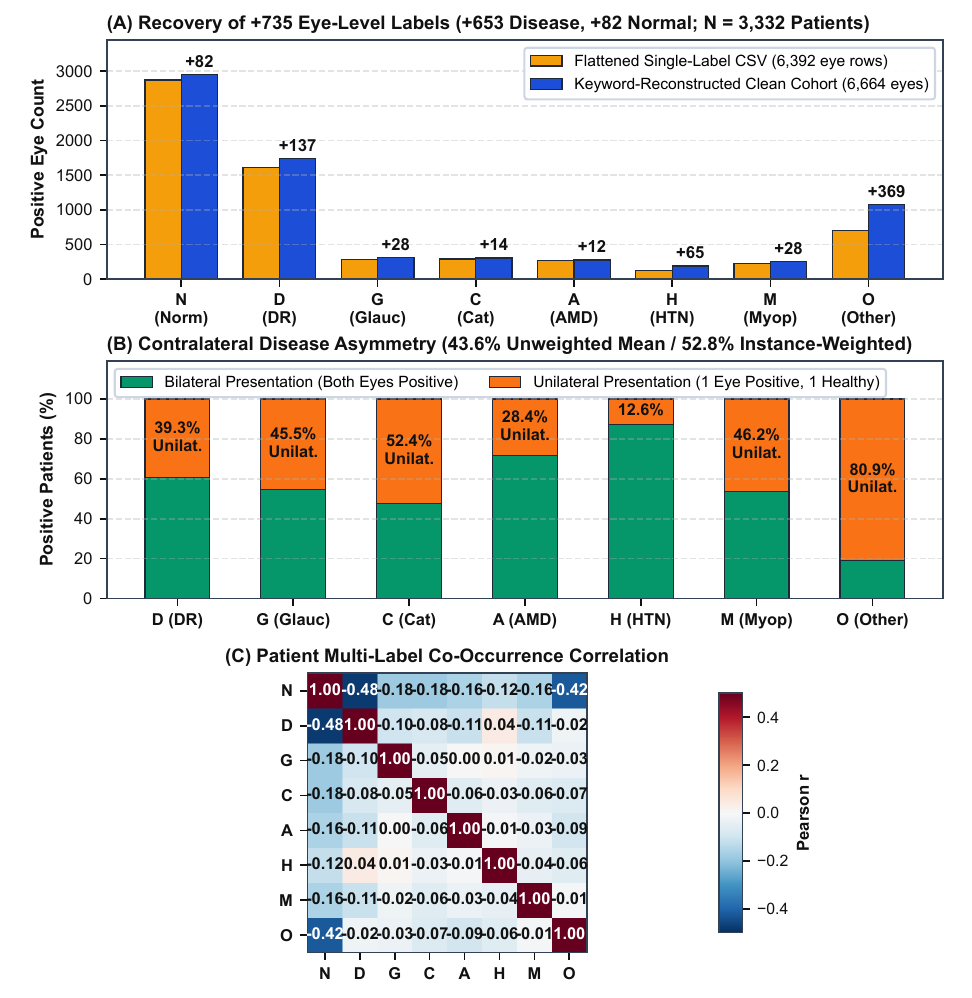}
\caption{Recovered labels and contralateral asymmetry in the retained gradable ODIR-5K cohort ($N=3{,}332$ patients, $6{,}664$ eyes). (A) Positive eye annotations gained over the flattened single-label CSV: a net $+735$ ($+653$ disease labels across \texttt{D--O} and $+82$ Normal \texttt{N} labels). (B) Bilateral versus unilateral disease breakdown across positive patients ($43.6\%$ unweighted mean across diseases; $52.8\%$ instance-weighted). (C) Multi-label co-occurrence correlation matrix.}\label{fig:dataset-audit}
\end{figure}

As summarized in Table~\ref{tab:dataset-flow} (Section~\ref{subsec:dataset-reconstruction}), deduplicating the source repository yields $3{,}358$ unique patients ($6{,}716$ eyes), and removing $26$ ungradable artifact-only pairs yields the retained gradable cohort of $3{,}332$ patients ($6{,}664$ eyes). Table~\ref{tab:dataset-audit} and Figure~\ref{fig:dataset-audit} detail the per-class eye and patient counts before and after keyword reconstruction. The flattened single-label CSV contains $6{,}392$ eye rows; it omits $324$ fellow eyes and truncates co-morbidities in multi-disease eyes. Within the retained $3{,}332$-patient gradable cohort summarized in Table~\ref{tab:dataset-audit} and plotted in Figure~\ref{fig:dataset-audit}A, keyword reconstruction restores a net $+735$ positive eye labels compared to the flattened single-label CSV, consisting of $+653$ disease labels (\texttt{D}: $+137$, \texttt{G}: $+28$, \texttt{C}: $+14$, \texttt{A}: $+12$, \texttt{H}: $+65$, \texttt{M}: $+28$, and \texttt{O}: $+369$) and $+82$ Normal (\texttt{N}) labels. In particular, positive eye instances for Hypertensive Retinopathy (\texttt{H}) increase by $50.8\%$ (from $128$ to $193$ eyes).

\begin{table}[!htbp]
\centering
\begin{widetab}[\widetabwidth]
\caption{Eye-level label recovery and contralateral disease laterality in the retained gradable ODIR-5K cohort ($N=3{,}332$ paired patients, $6{,}664$ eyes). CSV counts cover all $6{,}392$ rows of the flattened single-label file.}\label{tab:dataset-audit}
\scriptsize
\setlength{\tabcolsep}{4pt}
\renewcommand{\arraystretch}{1.15}
\begin{tabular*}{\linewidth}{@{\extracolsep{\fill}}llrrrrr@{}}
\toprule
\textbf{Class} & \textbf{Diagnostic Category} & \textbf{CSV Eyes} & \textbf{Clean Eyes} & \textbf{Recovered ($\Delta$)} & \textbf{Pos.\ Pat.} & \textbf{Unilateral (\%)} \\
\midrule
\textbf{N} & Normal Fundus & $2{,}873$ & $2{,}955$ & $+82$ ($+2.9\%$) & $1{,}079$ & $0$ (N/A) \\
\textbf{D} & Diabetic Retinopathy & $1{,}608$ & $1{,}745$ & $+137$ ($+8.5\%$) & $1{,}086$ & $427$ ($39.3\%$) \\
\textbf{G} & Glaucoma & $284$ & $312$ & $+28$ ($+9.9\%$) & $202$ & $92$ ($45.5\%$) \\
\textbf{C} & Cataract & $293$ & $307$ & $+14$ ($+4.8\%$) & $208$ & $109$ ($52.4\%$) \\
\textbf{A} & Age-Related Macular Deg. & $266$ & $278$ & $+12$ ($+4.5\%$) & $162$ & $46$ ($28.4\%$) \\
\textbf{H} & Hypertensive Retinopathy & $128$ & $193$ & $\mathbf{+65}$ ($\mathbf{+50.8\%}$) & $103$ & $13$ ($12.6\%$) \\
\textbf{M} & Pathological Myopia & $232$ & $260$ & $+28$ ($+12.1\%$) & $169$ & $78$ ($46.2\%$) \\
\textbf{O} & Other Abnormalities & $708$ & $1{,}077$ & $+369$ ($+52.1\%$) & $904$ & $731$ ($80.9\%$) \\
\midrule
\multicolumn{2}{@{}l}{\textbf{Total (All 8 / 7 Diseases)}} & $6{,}392$ / $3{,}519$ & $7{,}127$ / $4{,}172$ & $\mathbf{+735}$ / $\mathbf{+653}$ & $2{,}834^{\ast}$ & $1{,}496^{\ast}$ ($\mathbf{43.6\%}$ / $\mathbf{52.8\%}^{\ast}$) \\
\bottomrule
\end{tabular*}
\par\vspace{2pt}\noindent\parbox{\linewidth}{\scriptsize $^{\ast}$Totals across the 7 disease categories (\texttt{D--O}; bilateral cases $= \text{Pos.\ Pat.} - \text{Unilateral} = 1{,}338$); $43.6\%$ is the unweighted class mean and $52.8\%$ ($1{,}496/2{,}834$) is instance-weighted.}
\end{widetab}
\end{table}

Furthermore, Table~\ref{tab:dataset-audit} and Figure~\ref{fig:dataset-audit}B show that, on average across the seven disease categories (\texttt{D--O}), $\mathbf{43.6\%}$ of positive patients per disease class have the condition in only one eye ($\mathbf{52.8\%}$, or $1{,}496/2{,}834$, across all positive patient-disease instances, compared to $1{,}338$ bilateral instances). Unilateral presentation ranges from $12.6\%$ ($13/103$) in Hypertensive Retinopathy (\texttt{H}) and $28.4\%$ ($46/162$) in AMD (\texttt{A}) to $39.3\%$ ($427/1{,}086$) in Diabetic Retinopathy (\texttt{D}), $45.5\%$ ($92/202$) in Glaucoma (\texttt{G}), $46.2\%$ ($78/169$) in Pathological Myopia (\texttt{M}), $52.4\%$ ($109/208$) in Cataract (\texttt{C}), and $80.9\%$ ($731/904$) in Other Abnormalities (\texttt{O}). With unilateral disease this common, copying patient-level targets to both eyes (\texttt{A1}) labels many healthy eyes as diseased during monocular training.

\subsection{Eye-Specific Supervision and Cross-Architecture Replication}\label{subsec:results-supervision}
\begin{figure}[!htbp]
\centering
\includegraphics[width=\linewidth]{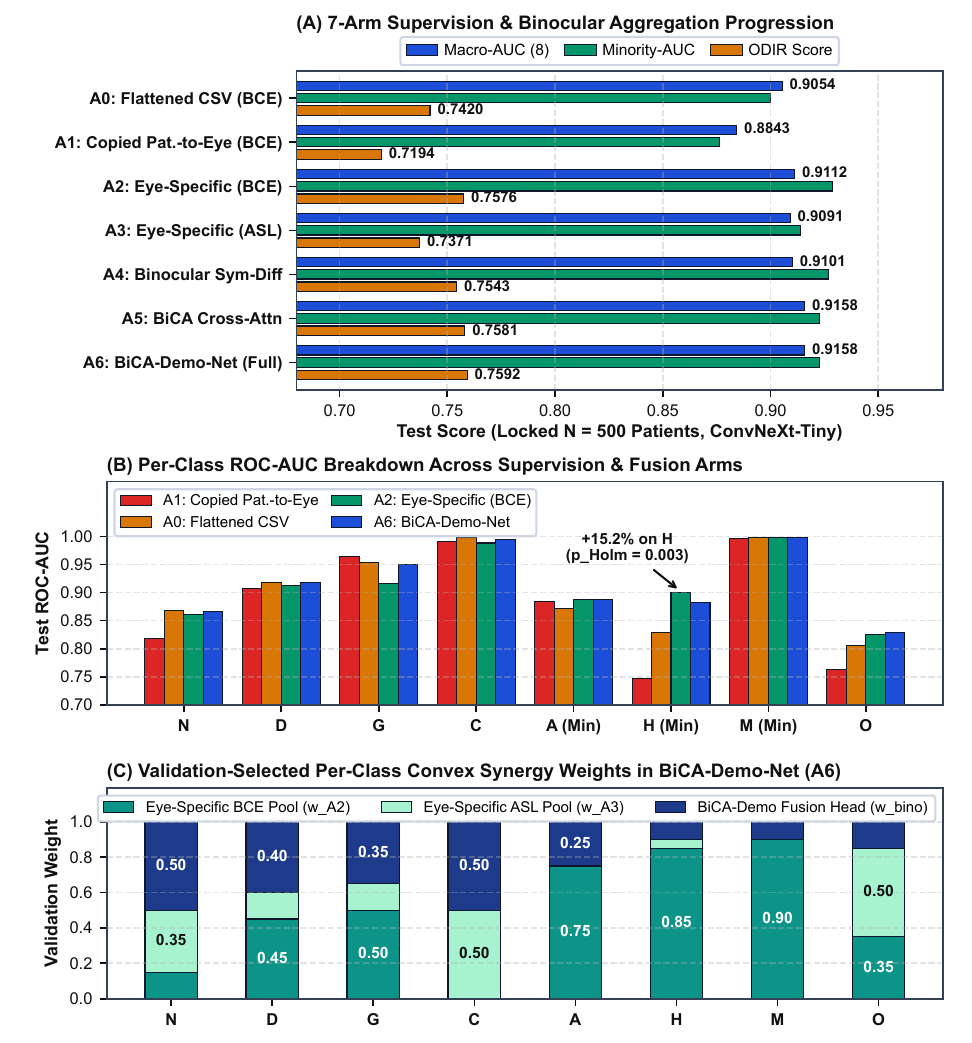}
\caption{Supervision and binocular aggregation ablation on the held-out test cohort ($N=500$ patients, \texttt{ConvNeXt-Tiny}). (A) Progression of Macro-AUC(8), Minority-AUC(A,H,M), and Official ODIR Score across arms \texttt{A0--A6}. (B) Per-class test ROC-AUC comparison. (C) Validation-selected per-class convex combination weights in \texttt{BiCA-Demo-Net} (\texttt{A6}).}\label{fig:ablation-progression}
\end{figure}

Table~\ref{tab:master-ablation} and Figure~\ref{fig:ablation-progression} report results of the 7-arm supervision and binocular fusion ablation (\texttt{A0}--\texttt{A6}) on \texttt{ConvNeXt-Tiny}, and cross-architecture replication on \texttt{ResNet-50d} on the locked $500$-patient test set ($1{,}000$ eyes). Because of the large number of summary metrics, to keep the presentation clear and readable, Panel~A of Table~\ref{tab:master-ablation} presents the primary summary metrics with $95\%$ patient-clustered bootstrap confidence intervals, whereas Panel~B presents the per-class ROC-AUC values across all eight categories ($\text{N, D, G, C, A, H, M,}$ and $\text{O}$). The paired patient-clustered bootstrap differences and permutation $p$-values are reported in Table~\ref{tab:bootstrap-contrasts}.

\begin{table}[!htbp]
\centering
\begin{widetab}[510pt]
\caption{Supervision and binocular fusion ablation across \texttt{ConvNeXt-Tiny} and \texttt{ResNet-50d} backbones on the locked patient-independent test set ($N=500$ patients, $1{,}000$ eyes). Panel A reports summary metrics with 95\% patient-clustered bootstrap confidence intervals ($2{,}000$ resamples); Panel B reports the per-class ROC-AUC breakdown.}\label{tab:master-ablation}
\scriptsize
\setlength{\tabcolsep}{3.5pt}
\renewcommand{\arraystretch}{1.15}
\begin{tabular*}{\linewidth}{@{\extracolsep{\fill}}lcccc@{}}
\toprule
\multicolumn{5}{@{}l}{\textbf{Panel A: Primary Summary Benchmark Metrics (Test Set, $N=500$ Patients)}} \\
\midrule
\textbf{Model Arm / Configuration} & \textbf{Macro-AUC [95\% CI]} & \textbf{Minority-AUC [95\% CI]} & \textbf{Macro-mAP} & \textbf{ODIR Score [95\% CI]} \\
\midrule
\multicolumn{5}{@{}l}{\emph{Primary Backbone: \texttt{ConvNeXt-Tiny} ($384\times 384$, $D=768$)}} \\
\texttt{A0}: Flattened CSV (BCE) & $0.9054$ $[0.8852, 0.9241]$ & $0.8999$ $[0.8502, 0.9425]$ & $\mathbf{0.7225}$ & $0.7420$ $[0.7142, 0.7675]$ \\
\texttt{A1}: Copied Pat.-to-Eye (BCE) & $0.8843$ $[0.8619, 0.9065]$ & $0.8764$ $[0.8227, 0.9232]$ & $0.6755$ & $0.7194$ $[0.6906, 0.7489]$ \\
\texttt{A2}: Eye-Specific (BCE) & $0.9112$ $[0.8918, 0.9289]$ & $\mathbf{0.9287}$ $[0.8887, 0.9634]$ & $0.7046$ & $0.7576$ $[0.7291, 0.7860]$ \\
\texttt{A3}: Eye-Specific (ASL) & $0.9091$ $[0.8882, 0.9271]$ & $0.9141$ $[0.8657, 0.9545]$ & $0.7096$ & $0.7371$ $[0.7118, 0.7626]$ \\
\texttt{A4}: Binocular Sym-Diff$^{\dagger}$ & $0.9101$ $[0.8902, 0.9279]$ & $0.9271$ $[0.8868, 0.9619]$ & $0.7036$ & $0.7543$ $[0.7275, 0.7806]$ \\
\texttt{A5}: BiCA Cross-Attn (No Demo)$^{\dagger}$ & $\mathbf{0.9158}$ $[0.8962, 0.9335]$ & $0.9226$ $[0.8782, 0.9613]$ & $0.7120$ & $0.7581$ $[0.7287, 0.7867]$ \\
\texttt{A6}: \texttt{BiCA-Demo-Net}$^{\dagger}$ & $\mathbf{0.9158}$ $[0.8966, 0.9332]$ & $0.9229$ $[0.8787, 0.9610]$ & $0.7120$ & $\mathbf{0.7592}$ $[0.7303, 0.7878]$ \\
\midrule
\multicolumn{5}{@{}l}{\emph{Cross-Architecture Replication: \texttt{ResNet-50d} ($384\times 384$, $D=2048$)}} \\
\texttt{R50d\_A1}: Copied Pat.-to-Eye (BCE) & $0.8932$ & $0.9150$ & $0.6585$ & $0.7085$ \\
\texttt{R50d\_A2}: Eye-Specific (BCE) & $0.9046$ & $0.9294$ & $0.6746$ & $0.7218$ \\
\texttt{R50d\_A3}: Eye-Specific (ASL) & $0.9033$ & $\mathbf{0.9388}$ & $\mathbf{0.6911}$ & $0.7156$ \\
\texttt{R50d\_A6}: \texttt{BiCA-Demo-Net}$^{\dagger}$ & $\mathbf{0.9099}$ & $0.9376$ & $0.6882$ & $\mathbf{0.7392}$ \\
\bottomrule
\end{tabular*}

\vspace{3pt}
\begin{tabular*}{\linewidth}{@{\extracolsep{\fill}}lcccccccc@{}}
\toprule
\multicolumn{9}{@{}l}{\textbf{Panel B: Per-Class ROC-AUC Breakdown Across All 8 Diagnostic Categories}} \\
\midrule
\textbf{Model Arm / Configuration} & \textbf{N} & \textbf{D} & \textbf{G} & \textbf{C} & \textbf{A} & \textbf{H} & \textbf{M} & \textbf{O} \\
\midrule
\texttt{A0}: Flattened CSV (BCE, ConvNeXt-T) & $0.8673$ & $\mathbf{0.9188}$ & $0.9539$ & $\mathbf{0.9983}$ & $0.8725$ & $0.8288$ & $0.9983$ & $0.8051$ \\
\texttt{A1}: Copied Pat.-to-Eye (BCE, ConvNeXt-T) & $0.8185$ & $0.9065$ & $\mathbf{0.9647}$ & $0.9918$ & $0.8846$ & $0.7475$ & $0.9972$ & $0.7633$ \\
\texttt{A2}: Eye-Specific (BCE, ConvNeXt-T) & $0.8602$ & $0.9128$ & $0.9168$ & $0.9885$ & $0.8879$ & $\mathbf{0.8995}$ & $\mathbf{0.9986}$ & $0.8256$ \\
\texttt{A3}: Eye-Specific (ASL, ConvNeXt-T) & $0.8548$ & $0.9086$ & $0.9624$ & $0.9929$ & $\mathbf{0.8966}$ & $0.8490$ & $0.9965$ & $0.8116$ \\
\texttt{A4}: Binocular Sym-Diff (ConvNeXt-T)$^{\dagger}$ & $0.8600$ & $0.9140$ & $0.9134$ & $0.9888$ & $0.8875$ & $0.8955$ & $0.9983$ & $0.8231$ \\
\texttt{A5}: BiCA Cross-Attn (No Demo, ConvNeXt-T)$^{\dagger}$ & $\mathbf{0.8693}$ & $0.9185$ & $0.9494$ & $0.9947$ & $0.8879$ & $0.8815$ & $0.9985$ & $0.8264$ \\
\texttt{A6}: \texttt{BiCA-Demo-Net} (ConvNeXt-T)$^{\dagger}$ & $0.8658$ & $0.9180$ & $0.9495$ & $0.9949$ & $0.8877$ & $0.8826$ & $0.9985$ & $\mathbf{0.8294}$ \\
\midrule
\texttt{R50d\_A1}: Copied Pat.-to-Eye (BCE, ResNet-50d) & $0.8122$ & $0.8852$ & $\mathbf{0.9229}$ & $0.9931$ & $0.9159$ & $0.8331$ & $\mathbf{0.9961}$ & $0.7871$ \\
\texttt{R50d\_A2}: Eye-Specific (BCE, ResNet-50d) & $\mathbf{0.8465}$ & $0.9010$ & $0.9182$ & $\mathbf{0.9935}$ & $0.9223$ & $0.8809$ & $0.9851$ & $0.7896$ \\
\texttt{R50d\_A3}: Eye-Specific (ASL, ResNet-50d) & $0.8280$ & $0.8917$ & $0.9049$ & $0.9921$ & $0.9210$ & $\mathbf{0.9037}$ & $0.9917$ & $0.7929$ \\
\texttt{R50d\_A6}: \texttt{BiCA-Demo-Net} (ResNet-50d)$^{\dagger}$ & $0.8444$ & $\mathbf{0.9029}$ & $0.9154$ & $0.9932$ & $\mathbf{0.9298}$ & $0.8914$ & $0.9917$ & $\mathbf{0.8102}$ \\
\bottomrule
\end{tabular*}
\par\vspace{2pt}\noindent\parbox{\linewidth}{\scriptsize $^{\dagger}$\texttt{A4} blends with \texttt{A2} (\texttt{BCE}) only ($w_{\text{A3}}=0$), whereas \texttt{A5} and \texttt{A6} (\texttt{BiCA-Demo-Net}) include the \texttt{A3} (\texttt{ASL}) pool (evaluated separately in Table~\ref{tab:fusion-comparison}).}
\end{widetab}
\end{table}

Three main findings emerge from Tables~\ref{tab:master-ablation} and \ref{tab:bootstrap-contrasts}:
\begin{enumerate}
    \item \textbf{Benefit of Eye-Specific Supervision (\texttt{A2} vs.\ \texttt{A1}):} Replacing copied patient-level labels (\texttt{A1}) with keyword-reconstructed eye-specific labels (\texttt{A2}) on \texttt{ConvNeXt-Tiny} increases 8-class Macro-AUC from $0.8843$ to $0.9112$ ($\Delta = +0.0270$, $p_{\text{raw}}=0.0100$), Minority-AUC from $0.8764$ to $0.9287$ ($\Delta = +0.0522$, $p_{\text{raw}}=0.0140$), and the official ODIR score from $0.7194$ to $0.7576$ (Table~\ref{tab:bootstrap-contrasts}). While the overall Macro-AUC and Minority-AUC $p$-values are borderline after 6-contrast Holm--Bonferroni correction ($p_{\text{Holm}}=0.0600$ and $0.0840$), the $+0.1521$ AUC improvement on Hypertensive Retinopathy (\texttt{H}: $0.7475 \to 0.8995$) remains significant after correction ($p_{\text{Holm}} = 0.0030$). On \texttt{ResNet-50d}, replacing copied patient labels (\texttt{R50d\_A1}: $0.8932$ Macro-AUC, $0.9150$ Minority-AUC, $0.8331$ Class-\texttt{H} AUC) with eye-specific labels shows the same overall direction of improvement: \texttt{R50d\_A2} (BCE) reaches $0.9046$ Macro-AUC, $0.9294$ Minority-AUC, and $0.8809$ Class-\texttt{H} AUC, while \texttt{R50d\_A3} (ASL) reaches $0.9033$ Macro-AUC, $0.9388$ Minority-AUC, and $0.9037$ Class-\texttt{H} AUC, although Glaucoma ($0.9229 \to 0.9182$) and Myopia ($0.9961 \to 0.9851$) decrease slightly under \texttt{R50d\_A2} (Table~\ref{tab:master-ablation} and Figure~\ref{fig:cross-backbone}A, B).
    \item \textbf{Recovery of Co-Morbid Hypertensive Retinopathy (\texttt{A2} vs.\ \texttt{A0}):} Because single-label truncation (\texttt{A0}) alters only multi-disease eyes, the overall 8-class Macro-AUC difference between \texttt{A2} and \texttt{A0} is small ($\Delta = +0.0059$, $p_{\text{raw}}=0.4163$). The picture differs for Hypertensive Retinopathy (\texttt{H}), which often accompanies Diabetic Retinopathy and lost $65$ eye labels in the flattened CSV. With multi-label targets restored, its AUC rises significantly, by $+0.0707$ ($0.8288 \to 0.8995$, $p_{\text{Holm}} = 0.0030$), and the official ODIR score moves from $0.7420$ to $0.7576$.
    \item \textbf{Complementary Behavior of BCE (\texttt{A2}) and ASL (\texttt{A3}):} Trained on identical eye-specific labels, the two losses are strong on different classes. \texttt{BCE} is better on Hypertensive Retinopathy (\texttt{H}: $0.8995$ vs.\ $0.8490$) and has the higher thresholded ODIR score ($0.7576$ vs.\ $0.7371$); \texttt{ASL} ranks Glaucoma (\texttt{G}: $0.9624$ vs.\ $0.9168$) and AMD (\texttt{A}: $0.8966$ vs.\ $0.8879$) better.
\end{enumerate}

\begin{table}[!htbp]
\centering
\begin{widetab}[510pt]
\caption{Patient-clustered bootstrap 95\% confidence intervals and Monte Carlo paired patient-clustered permutation $p$-values ($2{,}000$ resamples, with 6-contrast Holm--Bonferroni step-down adjustment) for key methodological contrasts on the locked test set ($N=500$ patients).}\label{tab:bootstrap-contrasts}
\scriptsize
\setlength{\tabcolsep}{4pt}
\renewcommand{\arraystretch}{1.15}
\begin{tabular*}{\linewidth}{@{\extracolsep{\fill}}lccc@{}}
\toprule
\textbf{Methodological Contrast} & \textbf{$\Delta$ Macro-AUC (8 Classes)} & \textbf{$\Delta$ Minority-AUC (A, H, M)} & \textbf{$\Delta$ Class-H AUC (Hypertension)} \\
\midrule
1. \texttt{A2} vs.\ \texttt{A1} (Copied Pat.) & \makecell[c]{$+0.0270$ $[+0.0089, +0.0457]$ \\ {\scriptsize $p=0.0100$ (Holm: $0.0600$)}} & \makecell[c]{$+0.0522$ $[+0.0100, +0.0943]$ \\ {\scriptsize $p=0.0140$ (Holm: $0.0840$)}} & \makecell[c]{$+0.1521$ $[+0.0470, +0.2671]$ \\ {\scriptsize $p<0.0005$ (\textbf{Holm: $\mathbf{0.0030}$})}} \\
2. \texttt{A2} vs.\ \texttt{A0} (Flattened CSV) & \makecell[c]{$+0.0059$ $[-0.0087, +0.0208]$ \\ {\scriptsize $p=0.4163$ (Holm: $1.0000$)}} & \makecell[c]{$+0.0288$ $[-0.0049, +0.0632]$ \\ {\scriptsize $p=0.0790$ (Holm: $0.3948$)}} & \makecell[c]{$+0.0707$ $[+0.0208, +0.1323]$ \\ {\scriptsize $p<0.0005$ (\textbf{Holm: $\mathbf{0.0030}$})}} \\
3. \texttt{A6} vs.\ \texttt{A1} (Copied Pat.) & \makecell[c]{$+0.0315$ $[+0.0130, +0.0504]$ \\ {\scriptsize $p=0.0285$ (Holm: $0.1424$)}} & \makecell[c]{$+0.0465$ $[+0.0031, +0.0907]$ \\ {\scriptsize $p=0.2164$ (Holm: $0.8656$)}} & \makecell[c]{$+0.1351$ $[+0.0183, +0.2592]$ \\ {\scriptsize $p=0.1839$ (Holm: $0.7356$)}} \\
4. \texttt{A6} vs.\ \texttt{A0} (Flattened CSV) & \makecell[c]{$+0.0104$ $[-0.0029, +0.0251]$ \\ {\scriptsize $p=0.4143$ (Holm: $1.0000$)}} & \makecell[c]{$+0.0231$ $[-0.0105, +0.0580]$ \\ {\scriptsize $p=0.3863$ (Holm: $1.0000$)}} & \makecell[c]{$+0.0538$ $[+0.0039, +0.1167]$ \\ {\scriptsize $p=0.4383$ (Holm: $1.0000$)}} \\
5. \texttt{A6} vs.\ \texttt{A2} (Eye-Specific) & \makecell[c]{$+0.0046$ $[-0.0016, +0.0108]$ \\ {\scriptsize $p=0.6507$ (Holm: $1.0000$)}} & \makecell[c]{$-0.0057$ $[-0.0167, +0.0019]$ \\ {\scriptsize $p=0.8501$ (Holm: $1.0000$)}} & \makecell[c]{$-0.0169$ $[-0.0496, +0.0058]$ \\ {\scriptsize $p=0.8926$ (Holm: $1.0000$)}} \\
6. \texttt{A6} vs.\ \texttt{A5} (Demo FiLM) & \makecell[c]{$+0.0001$ $[-0.0013, +0.0015]$ \\ {\scriptsize $p=0.9795$ (Holm: $1.0000$)}} & \makecell[c]{$+0.0003$ $[-0.0009, +0.0019]$ \\ {\scriptsize $p=0.9095$ (Holm: $1.0000$)}} & \makecell[c]{$+0.0012$ $[-0.0025, +0.0058]$ \\ {\scriptsize $p=0.7471$ (Holm: $1.0000$)}} \\
\bottomrule
\end{tabular*}
\end{widetab}
\end{table}

\subsection{Single-Backbone Binocular Fusion vs.\ Dual-Prior Logical Pooling}\label{subsec:fusion-results}
An important methodological question is whether the gain of \texttt{BiCA-\allowbreak Demo-Net} (\texttt{A6}: $0.9158$ Macro-AUC, $0.7592$ ODIR score) over monocular \texttt{A2} ($0.9112$ Macro-AUC, $0.7576$ ODIR score) stems from the learned cross-attention fusion head or from combining the complementary \texttt{A2} (BCE) and \texttt{A3} (ASL) monocular predictions. Table~\ref{tab:fusion-comparison} addresses this question through our 9-row binocular aggregation ablation, focusing on the summary benchmark metrics alongside the two diagnostic classes (\texttt{G} and \texttt{H}) that govern the BCE/ASL trade-off (complete 8-class values for \texttt{A2}--\texttt{A6} are given in Table~\ref{tab:master-ablation}, Panel~B).

\begin{table}[!htbp]
\centering
\begin{widetab}[510pt]
\caption{Nine-row binocular aggregation ablation comparing single-backbone pooling and learned fusion heads against dual-prior monocular pooling and \texttt{BiCA-Demo-Net} on the locked test set ($N=500$ patients, \texttt{ConvNeXt-Tiny}).}\label{tab:fusion-comparison}
\scriptsize
\setlength{\tabcolsep}{4pt}
\renewcommand{\arraystretch}{1.15}
\begin{tabular*}{\linewidth}{@{\extracolsep{\fill}}lcccccc@{}}
\toprule
\textbf{Binocular Aggregation Strategy} & \textbf{Class G} & \textbf{Class H} & \textbf{Macro-AUC} & \textbf{Min.-AUC} & \textbf{Macro-mAP} & \textbf{ODIR Score} \\
\midrule
1. Single-Backbone Mean-Pool: $0.5(p_L + p_R)$ & $0.9146$ & $\mathbf{0.9003}$ & $0.9090$ & $\mathbf{0.9309}$ & $0.6847$ & $0.7542$ \\
2. Single-Backbone Min/Max Pool (\texttt{A2} Only)$^{\ast}$ & $0.9166$ & $0.8997$ & $0.9112$ & $0.9287$ & $0.7046$ & $0.7587^{\ast}$ \\
3. Single-Backbone Concat MLP $[\bar{f}_L, \bar{f}_R]$ & $0.9112$ & $0.8997$ & $0.9105$ & $0.9286$ & $0.6974$ & $0.7576$ \\
4. Single-Backbone Sym-Diff $[\bar{f}_{\text{sym}}, |\bar{f}_L-\bar{f}_R|]$ & $0.9134$ & $0.8955$ & $0.9101$ & $0.9271$ & $0.7036$ & $0.7543$ \\
5. Single-Backbone BiCA Cross-Attn (No Demo) & $0.9154$ & $0.8972$ & $0.9104$ & $0.9279$ & $0.7024$ & $0.7452$ \\
6. Single-Backbone \texttt{BiCA-Demo-Net} (\texttt{A2} Only) & $0.9153$ & $0.8979$ & $0.9104$ & $0.9280$ & $0.7029$ & $0.7472$ \\
\midrule
7a. Fixed 50/50 Dual-Prior: $0.5(\text{A2}+\text{A3})^{\dagger}$ & $\mathbf{0.9581}$ & $0.8758$ & $\mathbf{0.9197}$ & $0.9277$ & $\mathbf{0.7218}$ & $0.7647$ \\
7b. Val-Opt Dual-Prior: $w\text{A2}+(1-w)\text{A3}^{\dagger}$ & $0.9491$ & $0.8861$ & $0.9170$ & $0.9242$ & $0.7128$ & $\mathbf{0.7666}$ \\
8. Dual-Prior \texttt{BiCA-Demo-Net} (\texttt{A6} Full) & $0.9495$ & $0.8826$ & $0.9158$ & $0.9229$ & $0.7120$ & $0.7592$ \\
\bottomrule
\end{tabular*}
\par\vspace{2pt}\noindent\parbox{\linewidth}{\scriptsize $^{\ast}$Evaluated on cached FP32 test logits ($\le 0.0002$ AUC and $0.0011$ ODIR-score difference due to validation F1 threshold discretization between FP16 and FP32 evaluation in Table~\ref{tab:master-ablation}). $^{\dagger}$Remaining class AUCs for Row 7a / Row 7b: $\text{N}=0.8703/0.8695$, $\text{D}=0.9211/0.9191$, $\text{C}=0.9946/0.9939$, $\text{A}=0.9097/0.8879$, $\text{M}=0.9976/0.9986$, $\text{O}=0.8301/0.8321$.}
\end{widetab}
\end{table}

Table~\ref{tab:fusion-comparison} highlights two practical findings for binocular retinal classification:
\begin{itemize}
    \item \textbf{Comparable Performance of Parameter-Free Min/Max Pooling on a Single Backbone:} Comparing Rows 1--6 shows that when a single backbone (\texttt{A2}) is trained on eye-specific targets, simple parameter-free min/max probability pooling ($\min(p_{L,N}, p_{R,N})$ and $\max(p_{L,c}, p_{R,c})$, Row 2: $0.9112$ Macro-AUC, $0.7046$ mAP, $0.7587$ ODIR score) performs as well as or slightly better than probability averaging (Row 1: $0.9090$), feature concatenation MLP (Row 3: $0.9105$), late symmetry-difference (Row 4: $0.9101$), and single-backbone cross-attention / \texttt{BiCA-Demo-Net} (Rows 5--6: $0.9104$). Because the ODIR-5K patient target follows the boolean rule in Eq.~\eqref{eq:logical-or}, parameter-free min/max pooling matches the annotation rule directly without adding extra trainable parameters that can overfit the $2{,}332$-patient training split.
    \item \textbf{Gain from Dual-Prior Pooling (Rows 7a \& 7b):} Combining the min/max pools of \texttt{A2} (BCE) and \texttt{A3} (ASL) without a neural fusion head yields the highest overall metrics in our study: fixed $50/50$ dual-prior pooling (Row 7a) reaches $\mathbf{0.9197}$ Macro-AUC, $\mathbf{0.7218}$ Macro-mAP, and $0.7647$ ODIR score, while validation-weighted dual-prior pooling (Row 7b) reaches the highest ODIR score of $\mathbf{0.7666}$ ($0.9170$ Macro-AUC). Both parameter-free dual-prior pools slightly exceed the full \texttt{BiCA-Demo-Net} configuration (Row 8: $0.9158$ Macro-AUC, $0.7592$ ODIR score). Thus, the $+0.0046$ Macro-AUC difference between \texttt{BiCA-Demo-Net} (\texttt{A6}) and \texttt{A2} ($p_{\text{raw}}=0.6507$, Table~\ref{tab:bootstrap-contrasts}) comes from combining the \texttt{BCE} and \texttt{ASL} models (which raises Glaucoma \texttt{G} AUC from $0.9166$ to $0.9495\text{--}0.9581$), while the \texttt{BiCA-Demo-Net} head primarily contributes spatial cross-attention visualizations (Figure~\ref{fig:gradcam}).
\end{itemize}

\subsection{Three-Seed Matched-Cohort Fellow-Eye Split Leakage Audit}\label{subsec:leakage-results}
\begin{figure}[!htbp]
\centering
\includegraphics[width=\linewidth]{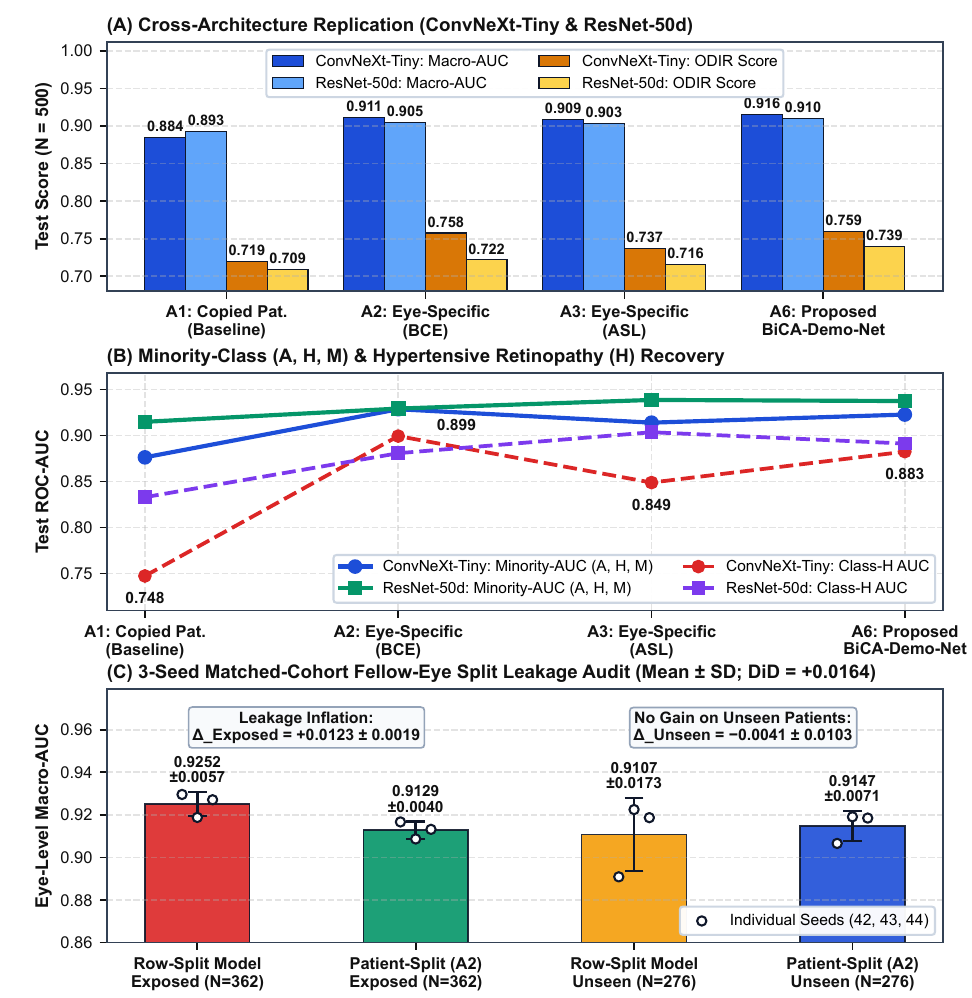}
\caption{Cross-architecture generalization and controlled fellow-eye split leakage audit. (A) Cross-backbone replication of Macro-AUC(8) and ODIR Score across \texttt{ConvNeXt-Tiny} and \texttt{ResNet-50d}. (B) Minority-class (\texttt{A, H, M}) and Class-\texttt{H} AUC trajectories across supervision arms. (C) Multi-seed matched-cohort comparison ($\text{Mean} \pm \text{SD}$ across seeds $42, 43, 44$, with individual seed markers) of Row-Split versus Patient-Split (\texttt{A2}) models on identical fellow-eye-exposed ($N=362$) and unseen-patient ($N=276$) test eyes.}\label{fig:cross-backbone}
\end{figure}

Table~\ref{tab:leakage-audit} and Figure~\ref{fig:cross-backbone}C report our 3-seed matched-cohort leakage experiment ($\text{seeds} \in \{42, 43, 44\}$), where an independently trained Row-Split Model and the Patient-Split Model (\texttt{A2}) are evaluated on the \emph{exact same} $362$ fellow-eye-exposed test eyes and the \emph{exact same} $276$ unseen-patient test eyes ($138$ patients).

\begin{table}[!htbp]
\centering
\begin{widetab}[\widetabwidth]
\caption{Multi-seed matched-cohort audit of fellow-eye split exposure comparing an independently trained Row-Split Model against the Patient-Split Model (\texttt{A2}) on identical held-out test eyes across three repeated stratified splits ($\text{seeds} \in \{42, 43, 44\}$).}\label{tab:leakage-audit}
\scriptsize
\setlength{\tabcolsep}{4pt}
\renewcommand{\arraystretch}{1.15}
\begin{tabular*}{\linewidth}{@{\extracolsep{\fill}}lccccccc@{}}
\toprule
 & \multicolumn{3}{c}{\textbf{Fellow-Eye-Exposed ($N=362$ Eyes)}} & \multicolumn{3}{c}{\textbf{Unseen-Patient ($N=276$ Eyes)}} & \textbf{Within-Seed} \\
\cmidrule(lr){2-4}\cmidrule(lr){5-7}
\textbf{Split Seed} & \textbf{Row-Split} & \textbf{Pat.\ (\texttt{A2})} & \textbf{$\Delta_{\text{Exposed}}$} & \textbf{Row-Split} & \textbf{Pat.\ (\texttt{A2})} & \textbf{$\Delta_{\text{Unseen}}$} & \textbf{$\text{DiD}$} \\
\midrule
Seed $42$ & $0.9296$ & $0.9167$ & $+0.0129$ & $0.8909$ & $0.9066$ & $-0.0158$ & $+0.0287$ \\
Seed $43$ & $0.9188$ & $0.9087$ & $+0.0101$ & $0.9225$ & $0.9191$ & $+0.0034$ & $+0.0067$ \\
Seed $44$ & $0.9271$ & $0.9132$ & $+0.0138$ & $0.9187$ & $0.9185$ & $+0.0001$ & $+0.0137$ \\
\midrule
\textbf{Mean} & $0.9252$ & $0.9129$ & $\mathbf{+0.0123}$ & $0.9107$ & $0.9147$ & $\mathbf{-0.0041}$ & $\mathbf{+0.0164}$ \\
\textbf{SD ($\pm$)} & $\pm 0.0057$ & $\pm 0.0040$ & $\mathbf{\pm 0.0019}$ & $\pm 0.0173$ & $\pm 0.0071$ & $\mathbf{\pm 0.0103}$ & $\mathbf{\pm 0.0112}$ \\
\bottomrule
\end{tabular*}
\par\vspace{2pt}\noindent\parbox{\linewidth}{\scriptsize \textbf{Seed 42 Permutation ($2{,}000$ swaps):} Exposed $\Delta = +0.0129$ ($95\%$ CI $[-0.0036, +0.0315]$, $p_{\text{perm}} = 0.1744$); Unseen $\Delta = -0.0158$ ($[-0.0385, +0.0042]$, $p_{\text{perm}} = 0.4048$). \textbf{Row-Split Class-H AUC (Seeds 42/43/44):} Exposed $= 0.9338 / 0.9109 / 0.9350$ ($0.9266 \pm 0.0136$); Unseen $= 0.9403 / 0.9415 / 0.8508$ ($0.9109 \pm 0.0520$).}
\end{widetab}
\end{table}

Across all three seeds, when the Row-Split Model is evaluated on test eyes whose fellow eye was in its training set ($N=362$), it scores higher than the Patient-Split Model (\texttt{A2}) on those same eyes by $+0.0129$, $+0.0101$, and $+0.0138$, giving a mean fellow-eye leakage inflation of $\mathbf{+0.0123 \pm 0.0019}$ Macro-AUC ($+1.23 \pm 0.19$ percentage points). In contrast, on the $276$ test eyes from patients with neither eye in training, the Row-Split Model's advantage disappears, averaging $\mathbf{-0.0041 \pm 0.0103}$ Macro-AUC relative to \texttt{A2}, with a mean Difference-in-Differences of $+0.0164 \pm 0.0112$ (Table~\ref{tab:leakage-audit}). Within a single seed of $362$ exposed eyes, the paired permutation contrast does not reach statistical significance on its own ($\Delta = +0.0129$, $p_{\text{perm}} = 0.1744$), so we treat this $+1.23\%$ shift as a consistent empirical inflation across seeds, whereas patient-label copying (\texttt{A1} vs.\ \texttt{A2}: $-2.70\%$ Macro-AUC, $-15.21\%$ Class-\texttt{H} AUC) is the larger source of performance loss.

\subsection{Platt Scaling and Validation-Locked Clinical Operating Points}\label{subsec:calibration-results}
\begin{figure}[!htbp]
\centering
\includegraphics[width=\linewidth]{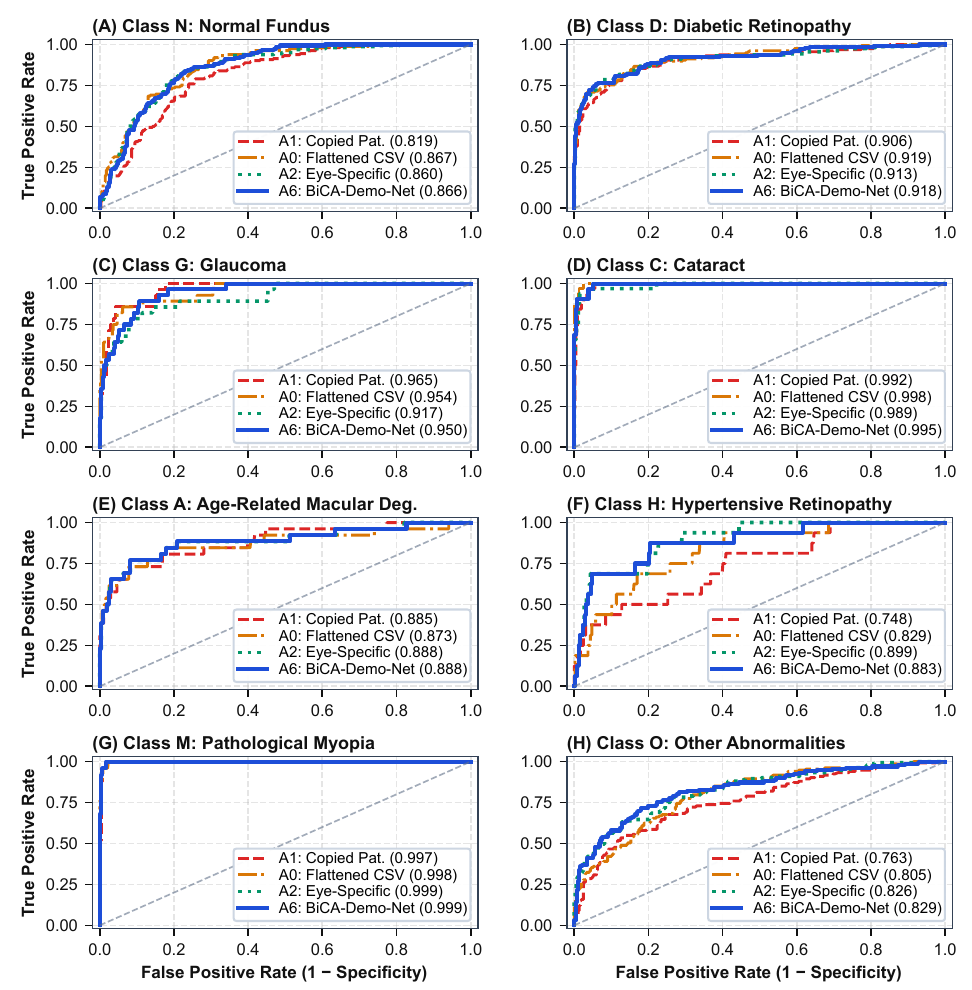}
\caption{Per-class Receiver Operating Characteristic (ROC) curves arranged in a $4\times 2$ grid across all 8 ophthalmic categories (\texttt{A}--\texttt{H} panels corresponding to classes \texttt{N, D, G, C, A, H, M, O}) on the locked patient-independent test cohort ($N=500$ patients), comparing \texttt{A1} (Copied Patient Label), \texttt{A0} (Flattened CSV), \texttt{A2} (Eye-Specific BCE), and \texttt{A6} (\texttt{BiCA-Demo-Net}).}\label{fig:roc-curves}
\end{figure}

\begin{figure}[!htbp]
\centering
\includegraphics[width=\linewidth]{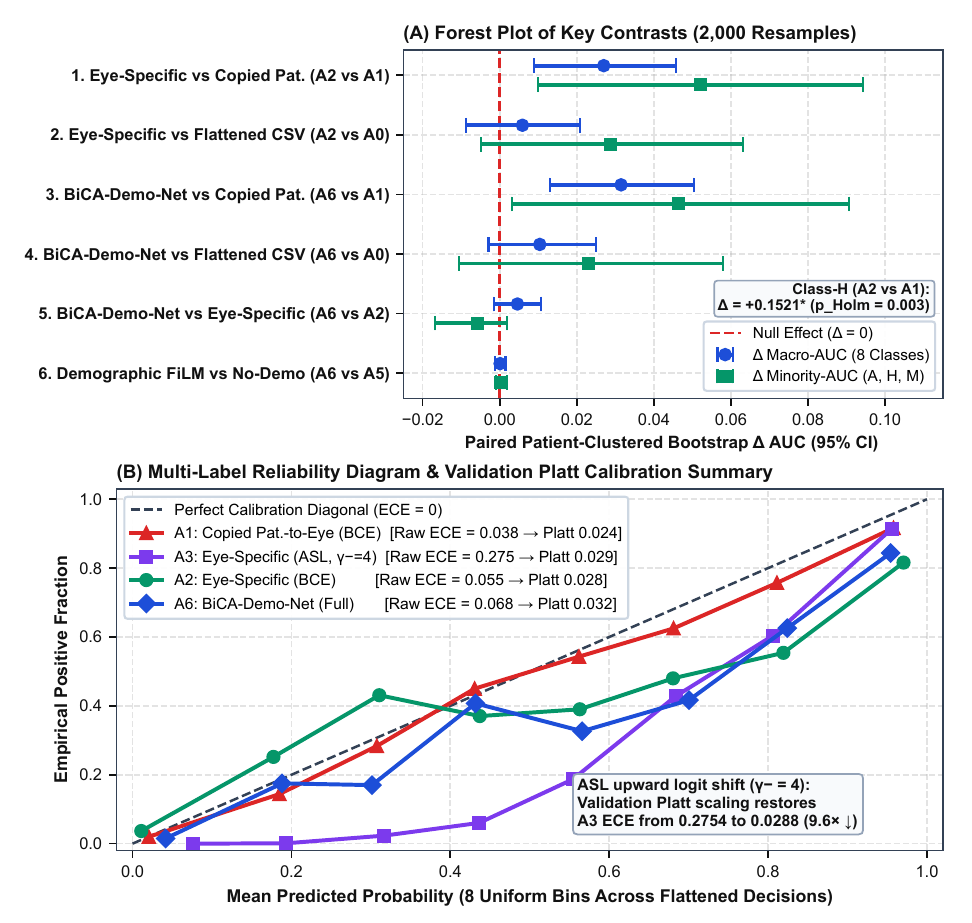}
\caption{Statistical significance and probability calibration analysis. (A) Forest plot of key methodological contrasts showing paired patient-clustered bootstrap $\Delta$ AUC with 95\% confidence intervals ($2{,}000$ resamples). (B) 8-bin uniform multi-label reliability calibration diagram across flattened decisions comparing raw predicted probabilities against empirical positive fractions, annotated with validation-fitted Platt-calibrated ECE reductions from Table~\ref{tab:calibration}.}\label{fig:forest-calibration}
\end{figure}

Table~\ref{tab:calibration} and Figures~\ref{fig:roc-curves}--\ref{fig:forest-calibration} report the probability calibration and clinical operating-point results on the $500$ held-out test patients (positive test patient counts per class: $n_N=162$, $n_D=158$, $n_G=28$, $n_C=32$, $n_A=26$, $n_H=16$, $n_M=25$, and $n_O=133$).

\begin{sidewaystable}[!htbp]
\centering
\caption{Raw and validation-fitted Platt-calibrated Brier score and Expected Calibration Error (ECE), alongside validation-locked clinical screening operating points on held-out test patients ($N=500$, \texttt{ConvNeXt-Tiny}; Class-\texttt{H} positive test patients $n_H = 16$).}\label{tab:calibration}
\scriptsize
\setlength{\tabcolsep}{3.5pt}
\renewcommand{\arraystretch}{1.15}
\begin{tabular*}{\textheight}{@{\extracolsep{\fill}}lcccccccc@{}}
\toprule
 & \multicolumn{2}{c}{\textbf{Raw Uncalibrated}} & \multicolumn{2}{c}{\textbf{Val.\ Platt-Calibrated}} & \multicolumn{4}{c}{\textbf{Val.-Locked Operating Points (Test Set)$^{\ast}$}} \\
\cmidrule(lr){2-3}\cmidrule(lr){4-5}\cmidrule(l){6-9}
\textbf{Model Arm} & \textbf{Brier $\downarrow$} & \textbf{ECE $\downarrow$} & \textbf{Brier $\downarrow$} & \textbf{ECE $\downarrow$} & \textbf{Macro Sens.} & \textbf{Min.\ Sens.} & \textbf{Class-H ($k/16$)} & \textbf{Spec@90Sens} \\
\midrule
\texttt{A0}: Flattened CSV (BCE) & $0.0684$ & $0.0540$ & $\mathbf{0.0614}$ & $0.0306$ & $71.91\%$ & $70.94\%$ & $43.75\%$ ($7/16$) & $73.99\%$ \\
\texttt{A1}: Copied Pat.-to-Eye (BCE) & $0.0696$ & $\mathbf{0.0382}$ & $0.0676$ & $\mathbf{0.0237}$ & $69.63\%$ & $68.86\%$ & $37.50\%$ ($6/16$) & $73.35\%$ \\
\texttt{A2}: Eye-Specific (BCE) & $0.0688$ & $0.0552$ & $\mathbf{0.0614}$ & $0.0278$ & $75.29\%$ & $\mathbf{77.99\%}$ & $\mathbf{68.75\%}$ ($\mathbf{11/16}$) & $78.88\%$ \\
\texttt{A3}: Eye-Specific (ASL) & $0.1548$ & $0.2754$ & $0.0636$ & $0.0288$ & $73.46\%$ & $69.93\%$ & $56.25\%$ ($9/16$) & $74.63\%$ \\
\texttt{A5}: BiCA Cross-Attn (No Demo) & $\mathbf{0.0675}$ & $0.0603$ & $0.0618$ & $0.0294$ & $77.35\%$ & $\mathbf{77.99\%}$ & $\mathbf{68.75\%}$ ($\mathbf{11/16}$) & $\mathbf{80.23\%}$ \\
\texttt{A6}: \texttt{BiCA-Demo-Net} & $0.0684$ & $0.0682$ & $0.0617$ & $0.0321$ & $\mathbf{77.45\%}$ & $\mathbf{77.99\%}$ & $\mathbf{68.75\%}$ ($\mathbf{11/16}$) & $79.51\%$ \\
\bottomrule
\end{tabular*}
\par\vspace{2pt}\noindent\parbox{\textheight}{\scriptsize $^{\ast}$At validation-locked 90\% specificity thresholds ($\tau_{90\text{Spec},c}$), achieved test specificities are $91.93\%$ (\texttt{A0}), $92.15\%$ (\texttt{A1}), $\mathbf{92.72\%}$ (\texttt{A2}), $91.85\%$ (\texttt{A3}), $92.17\%$ (\texttt{A5}), and $91.47\%$ (\texttt{A6}).}
\end{sidewaystable}

Two findings stand out in Table~\ref{tab:calibration}:
\begin{enumerate}
    \item \textbf{Effect of Platt Scaling on ASL Probability Calibration:} Although \texttt{A3} (\texttt{ASL}) achieves strong ranking performance ($0.9091$ Macro-AUC), its asymmetric negative down-weighting ($\gamma_- = 4$) yields a raw test ECE about five times higher than that of \texttt{A2} (\texttt{BCE}): $0.2754$ vs.\ $0.0552$, with a raw Brier score of $0.1548$ vs.\ $0.0688$ (Figure~\ref{fig:forest-calibration}B). Fitting per-class Platt scaling on the validation set reduces \texttt{A3}'s test ECE by $9.6$-fold (from $0.2754$ to $0.0288$) and its Brier score from $0.1548$ to $0.0636$, while also improving calibration for \texttt{A2} ($\text{ECE} = 0.0278$) and \texttt{BiCA-Demo-Net} (\texttt{A6}, $\text{ECE} = 0.0321$).
\end{enumerate}

\begin{enumerate}\setcounter{enumi}{1}
    \item \textbf{Sensitivity at Validation-Locked 90\% Specificity:} Locking the 90\% validation specificity thresholds produces achieved test specificities between $91.47\%$ and $92.72\%$ across all arms (Table~\ref{tab:calibration}). At this operating point, eye-specific supervision (\texttt{A2} and \texttt{BiCA-Demo-Net}) raises Macro sensitivity from $69.63\%$ (\texttt{A1}) to $75.29\%\text{--}77.45\%$, Minority-Class sensitivity from $68.86\%$ to $77.99\%$, and Hypertensive Retinopathy (\texttt{H}) sensitivity from $37.50\%$ ($6/16$ patients; $95\%$ Wilson CI $[18.5\%, 61.4\%]$) to $68.75\%$ ($11/16$ patients; $95\%$ Wilson CI $[44.4\%, 85.8\%]$). Only $n_H = 16$ test patients are positive for Class~\texttt{H}, so the binomial confidence intervals on this single-threshold gain are wide, and it is best read together with the increase in full ROC-AUC ($0.7475 \to 0.8995$, $p_{\text{Holm}}=0.0030$).
\end{enumerate}

\subsection{External Validation on APTOS-2019 and Qualitative Localization}\label{subsec:aptos-results}
To evaluate cross-dataset generalization without artificial duplicate-eye pairing, we apply our frozen monocular ODIR-5K classification heads directly to $N=3{,}662$ unpaired fundus photographs from the external APTOS-2019 blindness detection dataset \cite{aptos2019}, preprocessed with the same circular FOV cropping, square padding, and LAB-CLAHE ($384\times 384$) pipeline (Table~\ref{tab:aptos-external}).

\begin{table}[!htbp]
\centering
\begin{widetab}[\widetabwidth]
\caption{Zero-shot external transfer evaluation of monocular ODIR-trained classifiers on $N=3{,}662$ unpaired fundus images from the APTOS-2019 dataset under standardized FOV + LAB-CLAHE preprocessing ($1{,}000$-resample bootstrap 95\% CIs; $1{,}000$ resamples were used because the external cohort of $3{,}662$ images is more than seven times larger than the $500$-patient internal test set).}\label{tab:aptos-external}
\scriptsize
\setlength{\tabcolsep}{4pt}
\renewcommand{\arraystretch}{1.15}
\begin{tabular*}{\linewidth}{@{\extracolsep{\fill}}lcccc@{}}
\toprule
 & \multicolumn{2}{c}{\textbf{Any DR ($\text{Grade} \ge 1$, $n=1{,}857$)}} & \multicolumn{2}{c}{\textbf{Referable DR ($\text{Grade} \ge 2$, $n=1{,}487$)}} \\
\cmidrule(lr){2-3}\cmidrule(l){4-5}
\textbf{ODIR-5K Trained Model Arm} & \textbf{ROC-AUC [95\% CI]} & \textbf{AUPRC} & \textbf{ROC-AUC [95\% CI]} & \textbf{AUPRC} \\
\midrule
\multicolumn{5}{@{}l}{\emph{Primary Backbone: \texttt{ConvNeXt-Tiny} ($384\times 384$, $D=768$)}} \\
\texttt{A0}: Flattened Single-Label CSV (BCE) & $0.9753$ $[0.9701, 0.9803]$ & $0.9828$ & $0.9158$ $[0.9055, 0.9253]$ & $0.8204$ \\
\texttt{A1}: Copied Patient-to-Eye (BCE) & $0.9367$ $[0.9277, 0.9452]$ & $0.9574$ & $0.8721$ $[0.8593, 0.8856]$ & $0.7693$ \\
\texttt{A2}: Eye-Specific Supervision (BCE) & $0.9786$ $[0.9742, 0.9826]$ & $0.9823$ & $0.9149$ $[0.9055, 0.9240]$ & $0.8211$ \\
\texttt{A3}: Eye-Specific Supervision (ASL) & $0.9803$ $[0.9763, 0.9846]$ & $0.9852$ & $\mathbf{0.9375}$ $[0.9300, 0.9450]$ & $\mathbf{0.8890}$ \\
\textbf{Val-Weighted Dual-Prior ($w_{\text{A2}}=0.75$)} & $\mathbf{0.9848}$ $[0.9814, 0.9879]$ & $\mathbf{0.9875}$ & $0.9284$ $[0.9199, 0.9366]$ & $0.8598$ \\
\midrule
\multicolumn{5}{@{}l}{\emph{Cross-Architecture Replication: \texttt{ResNet-50d} ($384\times 384$, $D=2048$)}} \\
\texttt{R50d\_A1}: Copied Patient-to-Eye (BCE) & $0.9489$ $[0.9411, 0.9559]$ & $0.9624$ & $0.9019$ $[0.8915, 0.9117]$ & $0.8301$ \\
\texttt{R50d\_A2}: Eye-Specific Supervision (BCE) & $0.9712$ $[0.9660, 0.9763]$ & $0.9785$ & $0.9198$ $[0.9113, 0.9287]$ & $0.8524$ \\
\texttt{R50d\_A3}: Eye-Specific Supervision (ASL) & $\mathbf{0.9829}$ $[0.9787, 0.9870]$ & $\mathbf{0.9876}$ & $\mathbf{0.9274}$ $[0.9185, 0.9361]$ & $0.8561$ \\
\textbf{Val-Weighted Dual-Prior ($w_{\text{A2}}=0.75$)} & $0.9784$ $[0.9741, 0.9825]$ & $0.9834$ & $0.9254$ $[0.9171, 0.9336]$ & $\mathbf{0.8579}$ \\
\bottomrule
\end{tabular*}
\end{widetab}
\end{table}

Table~\ref{tab:aptos-external} confirms that models trained with copied patient-level labels (\texttt{A1}) transfer poorly to external diabetic retinopathy screening ($0.9367$ Any-DR AUC and $0.8721$ Referable-DR AUC on \texttt{ConvNeXt-Tiny}), likely because $39.3\%$ of Diabetic Retinopathy cases in ODIR-5K are unilateral (Table~\ref{tab:dataset-audit}). Training on eye-specific targets (\texttt{A2} and \texttt{A3}) raises \texttt{ConvNeXt-Tiny} Any-DR AUC to $0.9786\text{--}0.9803$ and Referable-DR AUC to $0.9149\text{--}0.9375$. The 95\% confidence intervals between \texttt{A1} and \texttt{A2}/\texttt{A3} are non-overlapping for \texttt{ConvNeXt-Tiny} on both tasks and for \texttt{ResNet-50d} on Any-DR, while slightly overlapping for \texttt{ResNet-50d} Referable-DR between \texttt{R50d\_A1} ($[0.8915, 0.9117]$) and \texttt{R50d\_A2} ($[0.9113, 0.9287]$) and non-overlapping for \texttt{R50d\_A3} ($[0.9185, 0.9361]$). Combining \texttt{A2} and \texttt{A3} via the ODIR validation weight ($w_{\text{A2}}=0.75$) reaches $0.9848$ Any-DR AUC on \texttt{ConvNeXt-Tiny}, while \texttt{R50d\_A3} reaches $0.9829$ Any-DR AUC on \texttt{ResNet-50d}.

\begin{figure}[!htbp]
\centering
\includegraphics[width=\linewidth]{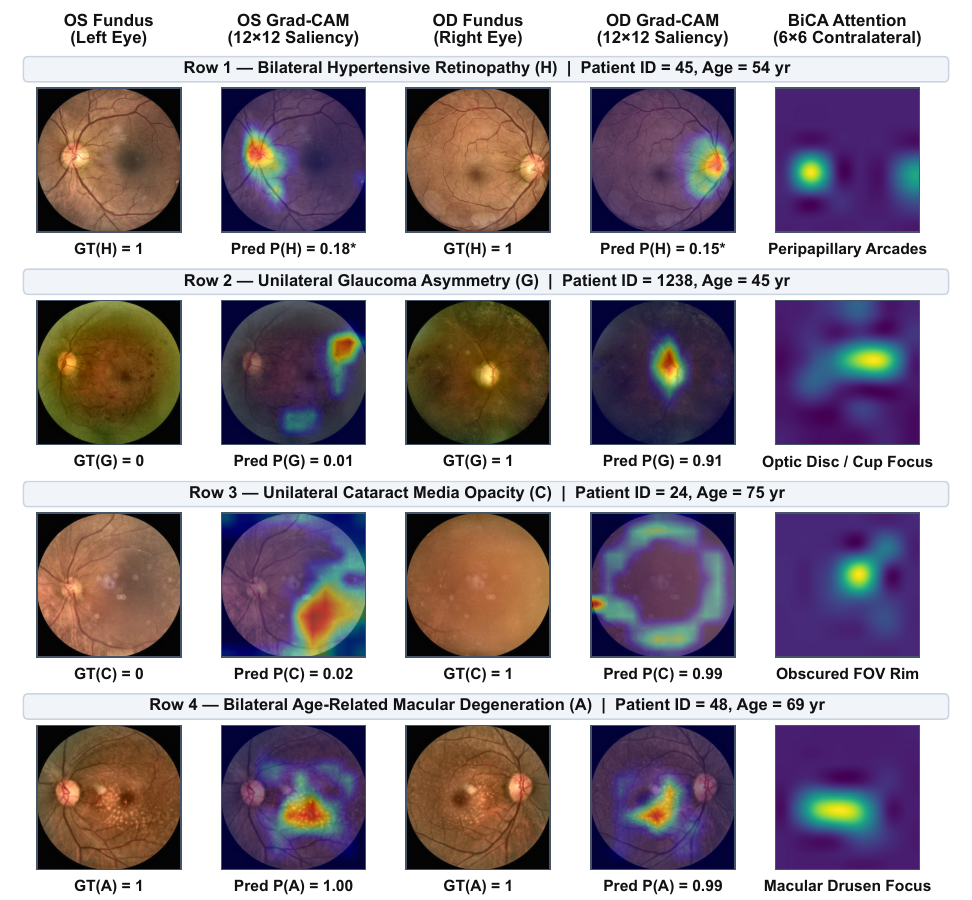}
\caption{Bilateral Grad-CAM saliency and $6\times 6$ contralateral spatial cross-attention maps displayed across uniform $1{:}1$ square tiles on four held-out test patients: (Row 1) Bilateral Hypertensive Retinopathy (\texttt{H}, Patient ID=45; raw \texttt{BCE} $P(\text{H})=0.18^{\ast}$ and $0.15^{\ast}$ reflect the $3.1\%$ base prevalence of Class \texttt{H}, so binary detection uses the class-specific validation threshold $\le 0.18$ rather than a naive $0.50$ cutoff), (Row 2) Unilateral Glaucoma asymmetry (\texttt{G}, Patient ID=1238), (Row 3) Unilateral Cataract (\texttt{C}, Patient ID=24, with activation concentrated along the obscured peripheral FOV rim), and (Row 4) Bilateral Age-Related Macular Degeneration (\texttt{A}, Patient ID=48).}\label{fig:gradcam}
\end{figure}

Finally, Figure~\ref{fig:gradcam} shows bilateral Grad-CAM \cite{selvaraju2017grad} saliency overlays and upsampled $6\times 6$ contralateral cross-attention maps on four held-out test patients. In Row 1 (Patient ID=45, bilateral Hypertensive Retinopathy \texttt{H}), the raw eye probabilities under \texttt{A2} (\texttt{BCE}) are $P(\text{H}) = 0.18$ (OS) and $P(\text{H}) = 0.15$ (OD). Because Class \texttt{H} has a base prevalence of only $3.1\%$ ($103/3{,}332$ patients), raw \texttt{BCE} probabilities for positive cases lie well below $0.50$: under a naive $0.50$ cutoff both eyes would be missed, whereas in our evaluation pipeline binary decisions use class-specific validation thresholds ($\tau_{\text{F1},\text{H}} \in [0.10, 0.90]$ in Table~\ref{tab:master-ablation} and $\tau_{90\text{Spec},\text{H}}$ in Table~\ref{tab:calibration}), so any validation threshold $\le 0.18$ classifies this patient as positive for Class \texttt{H}. Across the four examples, the $12\times 12$ Grad-CAM and $6\times 6$ cross-attention maps highlight peripapillary vascular arcades in bilateral Hypertensive Retinopathy (\texttt{H}, Row 1), the optic cup and disc in unilateral Glaucoma (\texttt{G}, Row 2), the hazily obscured peripheral circular FOV rim in unilateral Cataract (\texttt{C}, Row 3, where media opacity blurs the boundary rather than forming a focal central lesion), and macular drusen and pigmentary changes in bilateral AMD (\texttt{A}, Row 4).

\FloatBarrier
\subsection{Comparative Analysis with Published ODIR-5K Studies}\label{subsec:comparative-results}
Table~\ref{tab:literature-comparison} lists our proposed eye-specific framework and representative published ODIR-5K studies in a tabular form for a fair comparison in terms of split protocol, eye-level supervision source, backbone architecture, and multi-label evaluation metrics (Macro-AUC, F1-score, Cohen's $\kappa$, and Official ODIR Score; dashes denote measures not assessed in the original study).

\begin{table}[!htbp]
\centering
\begin{widetab}[\widetabwidth]
\caption{Comparative performance and methodological summary of representative published ODIR-5K studies versus our controlled baselines and proposed eye-specific framework. Bold marks the best value within Category~C; because Categories~A and~B use different split protocols and F1/$\kappa$ conventions, values are not directly comparable across categories.}\label{tab:literature-comparison}
\scriptsize
\setlength{\tabcolsep}{3.2pt}
\renewcommand{\arraystretch}{1.15}
\begin{tabularx}{\linewidth}{@{}l>{\raggedright\arraybackslash}Xlcccc@{}}
\toprule
\textbf{Study (Year)} & \textbf{Model / Architecture} & \textbf{Split \& Supervision} & \textbf{AUC} & \textbf{F1} & \textbf{$\kappa$} & \textbf{ODIR} \\
\midrule
\multicolumn{7}{@{}l}{\textbf{Category A: Published Patient-Split Studies (Copied Patient-Level Labels $Y_P$)}} \\
Wang et~al.\ (2020)~\cite{wang2020multi} & Dual EfficientNet-B3 Ensemble & Patient (9:1) / Copied $Y_P$ & $0.7400$ & $0.8800$ & $0.5800$ & $0.7300$ \\
Li et~al.\ (2021)~\cite{li2021benchmark} & Inception-v4 (Off-site test)$^{\dagger}$ & Patient split / Copied $Y_P$ & $0.8691$ & $0.8793$ & $0.5063$ & $0.7516$ \\
Li et~al.\ (2021)~\cite{li2021benchmark} & Inception-v4 (On-site test)$^{\dagger}$ & Patient split / Copied $Y_P$ & $0.8363$ & $0.8668$ & $0.4505$ & $0.7178$ \\
Gour \& Khanna (2021)~\cite{gour2021multiclass} & Two-Input VGG-16 CNN & Patient split / Copied $Y_P$ & $0.8493$ & $0.8557$ & --- & --- \\
He et~al.\ (2021)~\cite{he2021multi} & DCNet (ResNet-101 + Corr.) & Patient split / Copied $Y_P$ & $0.9300$ & $0.9130$ & --- & --- \\
Ou et~al.\ (2022)~\cite{ou2022bfenet} & BFENet (ResNet-50, On-site) & Patient split / Copied $Y_P$ & $0.9030$ & $0.8860$ & $0.5130$ & $0.7670$ \\
\midrule
\multicolumn{7}{@{}l}{\textbf{Category B: Published Flattened Single-Label CSV Studies (Eye-Row Evaluation)}} \\
Bhati et~al.\ (2023)~\cite{bhati2023discriminative} & DKCNet (Inception-ResNet-v2) & Eye-row split / Single CSV & $0.9608$ & $0.9428$ & $0.8100$ & --- \\
Wang et~al.\ (2023)~\cite{wang2023combining} & MBSaNet (CNN + Self-Attn.) & Eye-level / Single CSV & $0.9010$ & --- & $0.5040$ & $0.7490$ \\
Herrera-Chavez (2024)~\cite{herrera2024multilabel} & Transfer Learning CNNs & 5-fold row / Single CSV & $0.9450$ & --- & --- & --- \\
\midrule
\multicolumn{7}{@{}l}{\textbf{Category C: Controlled Baselines \& Proposed Framework (Strict Patient Split, $N=500$ Patients)}} \\
Baseline \texttt{A1} (Ours) & ConvNeXt-Tiny (BCE) & Strict patient / Copied $Y_P$ & $0.8843$ & $0.6654$ & $0.6085$ & $0.7194$ \\
Baseline \texttt{A0} (Ours) & ConvNeXt-Tiny (BCE) & Strict patient / Trunc.\ CSV & $0.9054$ & $0.6877$ & $0.6329$ & $0.7420$ \\
\textbf{Proposed \texttt{A2} (Ours)} & \textbf{ConvNeXt-Tiny + Min/Max Pool} & \textbf{Strict patient / 94-phrase} & $0.9112$ & $\mathbf{0.7056}$ & $0.6560$ & $0.7576$ \\
\textbf{Proposed \texttt{R50d\_A6} (Ours)} & \textbf{ResNet-50d + \texttt{BiCA-Demo-Net}} & \textbf{Strict patient / 94-phrase} & $0.9099$ & $0.6825$ & $0.6252$ & $0.7392$ \\
\textbf{Proposed \texttt{A6} (Ours)} & \textbf{ConvNeXt-Tiny + \texttt{BiCA-Demo-Net}} & \textbf{Strict patient / 94-phrase} & $0.9158$ & $0.7052$ & $\mathbf{0.6566}$ & $0.7592$ \\
\textbf{Proposed \texttt{Row 7a} (Ours)} & \textbf{ConvNeXt-Tiny 50/50 Dual-Prior} & \textbf{Strict patient / 94-phrase} & $\mathbf{0.9197}$ & --- & --- & $0.7647$ \\
\textbf{Proposed \texttt{Row 7b} (Ours)} & \textbf{ConvNeXt-Tiny Val-Opt Dual-Prior} & \textbf{Strict patient / 94-phrase} & $0.9170$ & --- & --- & $\mathbf{0.7666}$ \\
\bottomrule
\end{tabularx}
\par\vspace{2pt}\noindent\parbox{\linewidth}{\scriptsize $^{\dagger}$Li et~al.~\cite{li2021benchmark} report nine backbones under three feature-fusion rules; the rows above quote their element-wise multiplication (\texttt{PROD}) Inception-v4 results, which attain the highest ODIR score in their benchmark. Prior studies computing per-class binary F1 yield higher numerical scales than the flattened OIA-ODIR multi-label micro-F1/$\kappa$ protocol used in Category~C (Eq.~\eqref{eq:odir-score}).}
\end{widetab}
\end{table}

Key comparative observations from Table~\ref{tab:literature-comparison} are summarized below:
\begin{itemize}
    \item \textbf{Direct Head-to-Head Improvement (Category~C):} Under an identical backbone, resolution ($384\times 384$), and locked $500$-patient test split, replacing copied patient targets (\texttt{A1}) with 94-phrase eye-specific supervision improves Macro-AUC from $0.8843$ to $0.9197$ ($+3.54$ percentage points), Micro-F1 from $0.6654$ to $0.7056$, Cohen's $\kappa$ from $0.6085$ to $0.6566$, and Official ODIR Score from $0.7194$ to $0.7666$.
    \item \textbf{Comparison with Patient-Split Literature (Category~A):} Our eye-specific configurations outperform the OIA-ODIR Inception-v4 baselines~\cite{li2021benchmark}, Two-Input VGG-16~\cite{gour2021multiclass}, Dual EfficientNet-B3~\cite{wang2020multi}, and two-stream BFENet~\cite{ou2022bfenet} in Macro-AUC ($0.9197$ vs.\ $0.9030$) and Cohen's $\kappa$ ($0.6566$ vs.\ $0.5130$). Studies reporting higher internal AUCs~\cite{he2021multi} use custom internal split ratios or filtered subsets rather than the full $3{,}332$-pair multi-label cohort.
    \item \textbf{Accounting for Eye-Row CSV Studies (Category~B):} Higher raw AUCs ($0.9450\text{--}0.9608$) reported on the flattened single-label CSV~\cite{bhati2023discriminative,herrera2024multilabel} reflect single-label co-morbidity truncation combined with fellow-eye split exposure ($+1.23 \pm 0.19$ percentage-point inflation quantified in Table~\ref{tab:leakage-audit}).
\end{itemize}
\FloatBarrier

\section{Discussion}\label{sec:discussion}

The eye-specific supervision, data splitting, binocular aggregation, probability calibration, and external transfer are discussed in this section. It concludes with practical recommendations and limitations of this study.

\subsection{Eye-Specific Supervision}

When the patient label is copied to both eyes, a disease recorded in only one eye is also assigned to the other. In our retained cohort, an average of $43.6\%$ of positive patients across the seven disease categories had the condition in only one eye ($52.8\%$ across all positive patient-disease instances).

With \texttt{ConvNeXt-Tiny}, the macro-AUC was $0.9112$ with eye-specific supervision (\texttt{A2}) and $0.8843$ with patient-label copying (\texttt{A1}). This difference was not significant after Holm correction across six contrasts ($p_{\text{Holm}}=0.060$). For hypertensive retinopathy, the AUC increased from $0.7475$ to $0.8995$ ($p_{\text{Holm}}=0.003$). The gain did not appear in every class. \texttt{A1} had a higher AUC than \texttt{A2} for glaucoma ($0.9647$ versus $0.9168$) and cataract ($0.9918$ versus $0.9885$). In addition, hypertensive retinopathy had the lowest proportion of unilateral cases ($12.6\%$). Copying unilateral labels may therefore not be enough to explain its large gain, and our experiments do not identify the cause. On \texttt{ResNet-50d}, the macro-AUC was also higher with \texttt{A2} than with \texttt{A1} ($0.9046$ versus $0.8932$), although we did not test this difference statistically.

The difference between \texttt{A2} and simulated single-label truncation (\texttt{A0}) was smaller. Their macro-AUC values were $0.9112$ and $0.9054$, respectively (uncorrected $p=0.416$), and \texttt{A0} had the higher macro mean average precision ($0.7225$ versus $0.7046$). For hypertensive retinopathy, the AUC was higher with \texttt{A2} ($0.8995$ versus $0.8288$, $p_{\text{Holm}}=0.003$).

\subsection{Data Splitting and Fellow-Eye Exposure}

The matched-cohort audit compared the row-split and patient-split models on the same test eyes. For eyes whose fellow eyes were included in the training data of the row-split model, its macro-AUC was higher by $0.0123 \pm 0.0019$ across three seeds. For eyes from unseen patients, the mean difference was $-0.0041 \pm 0.0103$. The mean difference-in-differences was $0.0164 \pm 0.0112$. This trend is in line with a benefit of fellow-eye exposure. The difference on exposed eyes, however, was not significant in the permutation test with seed 42 ($p=0.174$), and the two-model training setups were not the same.

There are previous ODIR-5K studies that have reported higher AUC values compared to ours. These values cannot be directly compared due to the possibility of using different data splits, labels, cohorts, and metric definitions across studies. In some cases, part of the difference may be due to fellow-eye exposure, but we have not quantified the extent to which any gap is due to it.

\subsection{Binocular Aggregation}

In the labeling scheme, a disease is positive for a patient when it is positive in either eye, while the normal class follows a separate rule. Min/max pooling is motivated by this scheme and has no trainable parameters. With the \texttt{A2} backbone, it gave a macro-AUC of $0.9112$, and the other single-backbone methods ranged from $0.9090$ to $0.9105$. The ranking depended on the metric. For example, mean pooling had a slightly higher minority-class AUC than min/max pooling ($0.9309$ versus $0.9287$). We did not test whether the methods were statistically equivalent.

The full \texttt{BiCA-Demo-Net} system (\texttt{A6}) reached a macro-AUC of $0.9158$. Its difference from \texttt{A2} was not significant ($\Delta=+0.0046$, $95\%$ CI $[-0.0016, +0.0108]$, $p=0.651$). There was also not much difference between \texttt{A6} and \texttt{A5} (the system without demographic information) ($\Delta=+0.0001$, $p=0.980$). Since the \texttt{A6} output is a combination of the fusion head and the \texttt{A2} and \texttt{A3} predictions, this output does not show the head alone. The macro-AUC was $0.9197$ with a fixed equal combination of the \texttt{A2} and \texttt{A3} pooled predictions. Weighting the two predictions on the validation set gave the highest ODIR score of $0.7666$ in the aggregation comparison. Neither combination was statistically tested against \texttt{A2}. In this environment we could not see distinct evidence that the additional fusion layers outperformed more basic pooling. The attention maps were not evaluated quantitatively or by clinical specialists.

\subsection{Probability Calibration}

The Asymmetric Loss (\texttt{A3}) trained model had a macro-AUC of $0.9091$; however, the raw probabilities of the model were not well calibrated. Its expected calibration error (ECE) was $0.2754$, versus $0.0552$ for \texttt{A2}, which was trained using binary cross-entropy. This gap could be attributed to the different treatment of positive and negative samples in Asymmetric Loss. Platt scaling on the validation set decreased the ECE of \texttt{A3} to $0.0288$ and its Brier score from $0.1548$ to $0.0636$. After scaling, its ECE was close to that of \texttt{A2} ($0.0278$) but higher than that of \texttt{A1} ($0.0237$).

\subsection{External Transfer}

On APTOS-2019, the \texttt{ConvNeXt-Tiny} models trained with eye-specific labels had higher AUC values than the model trained with copied patient labels. For any diabetic retinopathy, the AUC was $0.9786$ for \texttt{A2}, $0.9803$ for \texttt{A3}, and $0.9367$ for \texttt{A1}. For referable diabetic retinopathy, the corresponding values were $0.9149$, $0.9375$, and $0.8721$. The advantage over \texttt{A0} was less consistent, and \texttt{A0} had a slightly higher AUC than \texttt{A2} when it came to referable diabetic retinopathy ($0.9158$ versus $0.9149$). This evaluation involved single images only and diabetic retinopathy alone. It did not assess binocular fusion or the other disease classes.

\subsection{Practical Recommendations}

The following points are useful to consider for future research with ODIR-5K. Instead of rows of images, patients are to be placed in the training, validation, and test sets, and patient identifiers are to be verified to avoid overlaps. The diagnostic keywords of each eye should be built into eye-level targets, not by copying the patient label onto each eye or by cutting each eye down to one label. The first baseline may be min/max pooling, and a mixture of binary cross-entropy and Asymmetric Loss models can be considered before fusion layers are implemented. In the case of Asymmetric Loss, Platt scaling needs to be fitted on the validation set, and the Brier score and ECE are to be reported together with ROC-AUC before clinical operating thresholds are chosen. Studies should also report the number of ungradable patient pairs dropped and the source image folders that were linked.

\subsection{Limitations and Future Work}\label{subsec:limitations}

There are several limitations to this study. Eye-level labels were created from the diagnostic text. There is good internal consistency of their agreement with the patient-level vectors, although the images were not independently regraded by ophthalmologists. Suspected glaucoma and suspected cataract were considered as positive cases.

The \texttt{A0} baseline simulated single-label truncation with a fixed priority order. It matched the original single-label column in $97.68\%$ of the training eyes present in the flattened file. In the case of hypertensive retinopathy, it retained $75$ positive training eyes, while the original column retained $87$ eyes. This might have enlarged a little the gap between \texttt{A0} and \texttt{A2} for this class.

For the main experiments, compute limits restricted the number of seeds to $1$. Evaluation was only conducted on \texttt{A1}, \texttt{A2}, \texttt{A3}, and \texttt{A6} for \texttt{ResNet-50d}. Out-of-fold features were not used; instead, features were cached from the training set and fed into the fusion heads. The results of the fusion heads might not be applicable to out-of-fold training due to a frozen backbone that can generate more confident predictions on the data it was trained on. The combination weights were selected for each class on the validation set, with a low number of positive samples in rare classes. Moreover, several aggregation approaches were evaluated on the test set and the best result of these should be interpreted with caution.

One backbone was used in the leakage audit. Its row-split models were trained for $12$ epochs, while the \texttt{A2} comparison model was trained for $15$ epochs. The left eye was used for training and the right eye for testing in the exposed patients. The difference-in-differences approach mitigates the impact of baseline differences, but not all impacts of the training setup.

The external evaluation only involved unpaired APTOS-2019 images. The sensitivity estimates for the hypertensive retinopathy class have wide confidence intervals in the internal test set, where only $16$ patients had this condition. In the future, a subset of discordant eye pairs should be independently graded, training should be rerun with multiple seeds, and out-of-fold training of the fusion heads should be performed. Label and leakage analyses should also be performed on other bilateral datasets as well as with retinal foundation models.

\section{Conclusion}\label{sec:conclusion}

In this study, we investigated the impact of label construction, data splitting, and binocular aggregation on multi-disease recognition on ODIR-5K, along with probability calibration. Under a patient-level split, the \texttt{ConvNeXt-Tiny} macro-AUC improved from $0.884$ to $0.911$ when moving from copied patient labels to eye-specific labels, although this difference was not statistically significant following Holm correction ($p_{\text{Holm}}=0.060$). The AUC for hypertensive retinopathy increased from $0.748$ to $0.900$ ($p_{\text{Holm}}=0.003$). In the matched-cohort audit, the row-split model had a macro-AUC higher by $0.012$ on test eyes whose fellow eyes were included in its training data, and the mean difference-in-differences was $0.016 \pm 0.011$. The full \texttt{BiCA-Demo-Net} system did not show a clear macro-AUC gain over min/max pooling of the \texttt{A2} model. Platt scaling reduced the ECE of the Asymmetric Loss model from $0.275$ to $0.029$. The main experiments used one seed, and the multi-disease findings come from ODIR-5K alone. They should next be tested with repeated training and on other bilateral datasets.
\section*{CRediT authorship contribution statement}
\textbf{Mohammed Ekram Ullah:} Conceptualization, Methodology, Software, Data curation, Formal analysis, Validation, Visualization, Writing -- original draft, Writing -- review \& editing.
\textbf{Abhijit Kumar Ghosh:} Validation, Writing -- review \& editing.
\textbf{Mohammad Zavid Parvez:} Supervision, Writing -- review \& editing.
\textbf{Subrata Chakraborty:} Supervision, Writing -- review \& editing.

\section*{Funding}
This research did not receive any specific grant from funding agencies in the public, commercial, or not-for-profit sectors.

\section*{Declaration of competing interest}
The authors declare that they have no known competing financial interests or personal relationships that could have appeared to influence the work reported in this paper.

\section*{Declaration of generative AI and AI-assisted technologies in the writing process}
During the preparation of this work, the authors used language-editing tools solely to improve grammar, readability, and \LaTeX{} formatting. After using these tools, the authors reviewed and edited the content as needed and take full responsibility for the content of the publication.

\section*{Data and code availability}
{\sloppy The source ODIR-5K and APTOS-2019 datasets are publicly available at \url{https://www.kaggle.com/datasets/andrewmvd/ocular-disease-recognition-odir5k} and \url{https://www.kaggle.com/datasets/mariaherrerot/aptos2019}. The reconstructed eye-level multi-label manifests (\texttt{odir5k\_\allowbreak binocular\_\allowbreak clean.csv} and \texttt{odir5k\_\allowbreak eye\_\allowbreak level\_\allowbreak clean.csv}), the 94-phrase taxonomy table (\texttt{phrase\_\allowbreak taxonomy\_\allowbreak audit.csv}), and the complete PyTorch reproduction code are provided in the supplementary material and public repository.\par}

\section*{Ethics statement}
This study exclusively analyzes fully de-identified, publicly released retrospective retinal fundus photographs from the OIA-ODIR and APTOS-2019 repositories; no new human subjects were recruited and institutional review board approval was not required.

\FloatBarrier
\appendix
\setcounter{table}{0}
\setcounter{figure}{0}
\renewcommand{\theHtable}{appendix.\thetable}
\renewcommand{\theHfigure}{appendix.\thefigure}
\section{Summary of the 94-Phrase Diagnostic Keyword Taxonomy and Reproducibility Mapping}\label{sec:appendix}

Table~\ref{tab:appendix-taxonomy} summarizes the mapping rules and phrase counts across the $94$ unique canonical (whitespace-normalized) diagnostic phrases identified across all $6{,}716$ eyes in ODIR-5K. The complete phrase-by-phrase frequency table is exported in \texttt{phrase\_\allowbreak taxonomy\_\allowbreak audit.csv}. For code-level reproducibility, the companion PyTorch notebook implements the dual-folder image path resolver as \texttt{resolve\_\allowbreak img\_\allowbreak path}, the 67-phrase ``Other Abnormality'' dictionary as \texttt{KNOWN\_\allowbreak O\_\allowbreak PHRASES}, the unmapped audit set as \texttt{UNKNOWN\_PHRASES}, the percentile-floored circular FOV ($\max(12, P_5(I_{\text{gray}}))$) and CIE-LAB CLAHE pipeline under the cache identifier \texttt{otsu\_\allowbreak fov\_\allowbreak sqpad\_\allowbreak lab\_\allowbreak clahe\_\allowbreak 384\_\allowbreak v4.1}, the pretrained \texttt{ConvNeXt-Tiny} checkpoint as \texttt{convnext\_tiny.\allowbreak fb\_in22k\_\allowbreak ft\_in1k\_384}, and the \texttt{BiCA-Demo-Net} fusion module as \texttt{FastBiCADemoFusionNet}.

\begin{table}[!htbp]
\centering
\begin{widetab}[\widetabwidth]
\caption{Summary of the deterministic 94-phrase diagnostic keyword taxonomy across all $6{,}716$ source eyes in ODIR-5K.}\label{tab:appendix-taxonomy}
\scriptsize
\setlength{\tabcolsep}{5pt}
\renewcommand{\arraystretch}{1.15}
\begin{tabularx}{\linewidth}{@{}llc>{\raggedright\arraybackslash}X@{}}
\toprule
\textbf{Target Category} & \textbf{ICD-10 Code} & \textbf{Phrases} & \textbf{Representative Canonical Phrases in Registry} \\
\midrule
Artifact / Quality (Excluded) & Non-diagnostic & $6$ & \texttt{lens dust}, \texttt{low image quality}, \texttt{optic disk photographically invisible}, \texttt{anterior segment image}, \texttt{image offset}, \texttt{no fundus image} \\
\textbf{N} (Normal Fundus) & Z01.00 & $1$ & \texttt{normal fundus} (assigned $N=1$ iff $0$ disease phrases present in eye) \\
\textbf{D} (Diabetic Retinopathy) & H36.0 / E11.3 & $9$ & \texttt{moderate non proliferative retinopathy}, \texttt{mild nonproliferative retinopathy}, \texttt{severe nonproliferative retinopathy}, \texttt{diabetic retinopathy}, \texttt{proliferative diabetic retinopathy} \\
\textbf{G} (Glaucoma) & H40.0--H40.9 & $2$ & \texttt{glaucoma}, \texttt{suspected glaucoma} \\
\textbf{C} (Cataract) & H25--H26 & $2$ & \texttt{cataract}, \texttt{suspected cataract} \\
\textbf{A} (AMD) & H35.3 & $3$ & \texttt{dry age-related macular degeneration}, \texttt{wet age-related macular degeneration}, \texttt{age-related macular degeneration} \\
\textbf{H} (Hypertensive Retinopathy) & H35.0 & $1$ & \texttt{hypertensive retinopathy} (the sole Class-\texttt{H} phrase in ODIR-5K; no severity-graded variants occur) \\
\textbf{M} (Pathological Myopia) & H44.2 & $3$ & \texttt{pathological myopia}, \texttt{myopia retinopathy}, \texttt{myopic maculopathy} \\
\textbf{O} (Verified Other Abnormality) & H30--H35, H47 & $67$ & \texttt{drusen}, \texttt{epiretinal membrane}, \texttt{macular hole}, \texttt{branch retinal vein occlusion}, \texttt{central retinal vein occlusion}, \texttt{tessellated fundus} \\
\midrule
\textbf{Unmapped / Unknown Set} & --- & $\mathbf{0}$ & Verified empty ($|\texttt{UNKNOWN\_PHRASES}| = 0$) across all $6{,}716$ eyes \\
\bottomrule
\end{tabularx}
\end{widetab}
\end{table}

\FloatBarrier
\bibliographystyle{elsarticle-num}
\bibliography{references}

\end{document}